\documentclass{article}
\usepackage[T1]{fontenc}
\usepackage{microtype,graphicx,booktabs,multirow,array,tabularx}
\usepackage{hyperref}
\usepackage{xurl}
\usepackage[preprint]{icml2026}
\usepackage{amsmath,amssymb,mathtools,amsthm}
\usepackage[capitalize,noabbrev]{cleveref}
\usepackage{placeins}
\usepackage{enumitem}

\graphicspath{{assets/}}
\newcommand{\R}{\mathbb R}
\newcommand{\E}{\mathbb E}
\newcommand{\sg}{\operatorname{sg}}
\newcommand{\LN}{\operatorname{LN}}
\newcommand{\tbd}[1]{\par\noindent\textbf{Pending evidence.} #1\par}
\newcolumntype{Y}{>{\raggedright\arraybackslash}X}
\icmltitlerunning{Re-JEPA: Coupling Representation Prediction and Signal Reconstruction for State-Conditioned EMG Tracking}
\icmlkeywords{self-supervised learning, EMG, contextual prediction, signal reconstruction, hand motion tracking}
\hypersetup{pdftitle={Re-JEPA: Coupling Representation Prediction and Signal Reconstruction for State-Conditioned EMG Tracking}}
\begin{document}
\twocolumn[
\icmltitle{Re-JEPA: Coupling Representation Prediction and Signal Reconstruction for State-Conditioned EMG Tracking}
\icmlsetsymbol{cor}{\textdagger}
\icmlsetsymbol{lead}{*}
\begin{icmlauthorlist}
\icmlauthor{Yijian Yang}{cas,comp}
\icmlauthor{Tao Chen}{cor,lead,comp}
\icmlauthor{Xiaodong Yang}{cor,cas}
\icmlauthor{Dalong Du}{comp,tsu}
\icmlauthor{Zongheng Guo}{polimi,comp}
\icmlauthor{Heng Liu}{comp}
\icmlauthor{Lun SHU}{comp}
\icmlauthor{Chen Luo}{zju,comp}
\icmlauthor{Junhao Tu}{comp}
\icmlauthor{Lei Shu}{bit,comp}
\icmlauthor{Mingzhou Wu}{comp}
\icmlauthor{Yifan Liu}{comp}
\icmlauthor{Zhuanghui Zhang}{comp}
\icmlauthor{Zhenqiang Liu}{comp}
\icmlauthor{Shilei Wen}{comp}
\end{icmlauthorlist}
\icmlaffiliation{polimi}{Department of Electronics, Information and Bioengineering, Politecnico di Milano, Milan, Italy}
\icmlaffiliation{comp}{SynapX, Beijing, China}
\icmlaffiliation{tsu}{Department of Automation, Tsinghua University, Beijing, China}
\icmlaffiliation{zju}{State Key Laboratory of Industrial Control Technology, Zhejiang University, Hangzhou, China}
\icmlaffiliation{cas}{Institute of Computing Technology, Chinese Academy of Sciences, Beijing, China}
\icmlaffiliation{bit}{School of Automation, Beijing Institute of Technology, Beijing, China}
\icmlcorrespondingauthor{Tao Chen}{shubing@synapa.ai}
\icmlcorrespondingauthor{Xiaodong Yang}{yangxiaodong@ict.ac.cn}
\vskip 0.3in
]
\printAffiliationsAndNotice{\textsuperscript{\textdagger}Corresponding authors. \textsuperscript{*}Project leader. }
\hypersetup{pdfsubject={Preprint}}
\hypersetup{pdftitle={Re-JEPA: Coupling Representation Prediction and Signal Reconstruction for State-Conditioned EMG Tracking}}
\begin{abstract}
Improving continuous hand tracking from surface electromyography (sEMG) requires generalization across users and motion types while maintaining sensitivity to subtle movements. Joint-embedding predictive architectures (JEPAs) learn latent representation spaces through contextual prediction, but their effectiveness for tracking depends on which motion-relevant details these representations preserve and how those details are interpreted as the hand state evolves. We introduce Re-JEPA, a tracking framework that couples representation prediction with signal reconstruction and incorporates state-conditioned motion decoding. During pretraining without kinematic labels, Re-JEPA predicts continuous latent targets at multiple levels from visible EMG patches and reconstructs masked amplitude and phase attributes from the fused predictions. Signal reconstruction imposes an additional constraint on the contextual predictor, encouraging its outputs to retain information needed for signal recovery. For downstream tracking, the Gated Persistent Motion (GPM) decoder conditions EMG-token readout on recurrent state tokens and motion history, then generates continuous trajectories through gated state updates and incremental pose prediction. Under the emg2pose tracking protocol, Re-JEPA reduces angular MAE after full fine-tuning by approximately 10.3\%, 19.3\%, and 14.9\% relative to vemg2pose on the cross-user, cross-stage, and joint cross-user–stage splits, respectively.
\end{abstract}

\section{Introduction}
\label{sec:intro}
\begin{figure*}[!t]
\centering
\includegraphics[width=\textwidth]{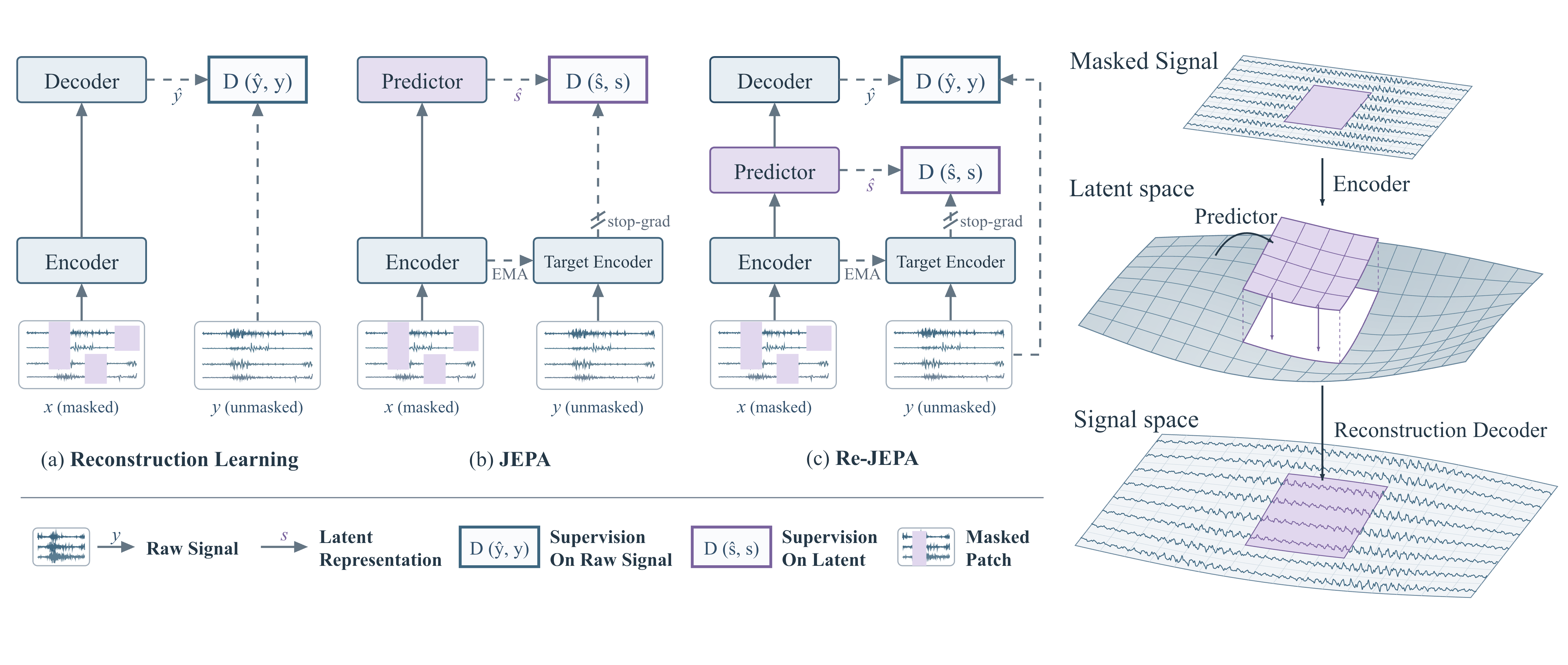}
\caption{Three self-supervised paths and their supervision spaces. Left: (a) reconstruction learning supervises the recovery of masked observations in signal space; (b) JEPA matches continuous target representations computed from the complete input; (c) Re-JEPA reconstructs after prediction, propagating recovery gradients through the predictor and online encoder. The masked and complete signal inputs are denoted by $x$ and $y$, respectively. The target encoder is updated by EMA, and its outputs are stop-gradient targets. Right: missing representations are predicted in representation space and decoded into signal attributes. Light purple marks masked or to-be-recovered regions. The surface is a conceptual illustration, not a measured embedding manifold.}
\label{fig:paradigms}
\end{figure*}

Continuous hand trajectories provide human demonstrations for learning dexterous manipulation in embodied systems \citep{egodex}. Egocentric visual capture, however, is limited by hand--object occlusion and field of view. Surface electromyography (sEMG) records electrical activity associated with muscle activation, offering an alternative source for continuous hand tracking without requiring visual visibility. EMG is not a direct measurement of hand pose: interpreting its movement-related content requires temporal context, while differences in users, sensor placement, and motion patterns complicate cross-user and cross-stage generalization \citep{emg2pose}. Tracking representations must therefore use context to interpret muscle activity while preserving signal differences associated with subtle motion updates. Self-supervised pretraining on EMG without kinematic annotations offers a way to learn such representations.

Physiological foundation models demonstrate the potential of learning transferable representations from large unlabeled datasets. LaBraM, CBraMod, CSBrain, and SIGMA-PPG use discrete-code prediction or signal reconstruction for physiological signal pretraining \citep{labram,cbramod,csbrain,guo2026sigma}. Reconstruction supervises recoverable detail, but accurate recovery does not ensure that the information is organized for downstream readout; noisy targets may also encourage recovery of predictable, task-irrelevant fluctuations. LeCun and colleagues have explored joint-embedding predictive architectures (JEPA), which estimate target-region representations from visible context. Using learned abstract representations instead of observations as prediction targets allows learning to emphasize structured and semantic relationships without recovering every low-level detail. I-JEPA implements this idea for images without relying on handcrafted view augmentations; V-JEPA extends feature prediction to video, and EEG-JEPA introduces multilevel contextual prediction for physiological signals \citep{ijepa,vjepa,eegjepa}. These developments motivate EMG pretraining for transferable contextual relationships. Continuous tracking, however, also depends on small observational differences that may encode motion updates. Representation matching alone does not directly require every such difference to be retained. We therefore seek to preserve the benefits of contextual prediction while constraining its intermediate representations to retain fine-grained signal information.

We propose \textbf{Re-JEPA}, an EMG pretraining framework that first predicts in representation space, then decodes the predictions into signal space under a recovery constraint (\cref{fig:paradigms}). The model predicts the continuous representations of a momentum target encoder at multiple depths from visible patches. It fuses these predictions, restores their positions in a complete token grid, and uses amplitude and phase decoders to recover the signal attributes of target regions. Reconstruction gradients propagate through the predictor to the online encoder, exposing the same intermediate predictions to representation matching and signal recovery. This path incorporates signal recoverability into contextual representation learning, retaining information that continuous tracking may require. Overlapping patches refine the temporal grid, while masking physical regions in the original signal controls the visibility of target content.

Using these pretrained tokens for continuous tracking further requires interpreting EMG observations in light of motion history. We introduce the \textbf{Gated Persistent Motion (GPM)} decoder, which maintains recurrent state tokens and explicitly conditions on the previous pose estimate and motion increment. GPM reads EMG tokens through attention, updates its latent state through a gate, and predicts the next pose increment for accumulation. It connects context-dependent signal representations to state-conditioned motion decoding. Frozen readouts, full fine-tuning, and pretraining-objective comparisons evaluate continuous tracking and cross-user and cross-stage generalization.

Our contributions are threefold:
\begin{enumerate}[topsep=4pt,partopsep=0pt,itemsep=3pt,parsep=0pt]
\item \textbf{Re-JEPA pretraining for continuous EMG tracking.} Amplitude and phase recovery follows multilevel continuous representation prediction, jointly optimizing contextual matching and signal recoverability along the same computational path.
\item \textbf{State-conditioned GPM decoding.} Recurrent state tokens, explicit motion conditions, and EMG-token attention enable gated memory updates and incremental pose decoding.
\item \textbf{Improved tracking across users and stages.} Under the emg2pose protocol with a provided initial pose and user-balanced aggregation, angular MAEs are 6.91$^\circ$, 9.04$^\circ$, and 9.36$^\circ$ on the three generalization splits, approximately 10.3\%, 19.3\%, and 14.9\% below the published vemg2pose means \citep{emg2pose}. With the same Fusion+GPM96 downstream interface and test protocol, joint pretraining reduces fine-tuning angular errors by 4.6\%, 8.4\%, and 7.0\% relative to latent-only pretraining.
\end{enumerate}

\section{Related Work}
\subsection{EMG-Based Continuous Hand Motion Tracking}
Continuous hand tracking from sEMG estimates time-varying joint poses rather than discrete gestures. NeuroPose uses wearable EMG for 3D hand pose tracking, while SensingDynamics models continuous hand kinematics from high-density sEMG \citep{neuropose,sensingdynamics}. The emg2pose benchmark provides large-scale synchronized EMG and kinematics, cross-user, cross-stage, and joint generalization splits, and a distinction between tracking with an initial pose and regression without one \citep{emg2pose}. Subsequent work studies decoding parameterization, user adaptation, and physiological inductive biases. Comparisons of position and velocity decoding reveal the influence of output scaling and training settings; REACT infers conditioning representations from a small amount of user calibration data to modulate shared features; NHN combines spatiotemporal encoding with physiologically motivated latent neuromotor states \citep{positionvelocity,react,nhn}.

Beyond supervised decoding, recent studies use EMG without kinematic annotations to learn transferable representations. EMG-GPT constructs discrete EMG codes with a frozen residual vector-quantization tokenizer, predicts future codes through causal temporal modeling, and transfers the resulting features to tracking and regression \citep{emggpt}. In contrast to prediction with frozen discrete targets, Re-JEPA predicts multilevel continuous representations of masked regions from visible context and applies amplitude and phase recovery constraints to those predictions. GPM then reads the pretrained tokens using recurrent motion state to update the pose.

\subsection{Masked Reconstruction for Time-Series Pretraining}
Masked reconstruction learns time-series representations by recovering hidden observations from visible context. PatchTST segments time series into patches, randomly masks non-overlapping segments during self-supervised pretraining, and reconstructs their values with mean squared error \citep{patchtst}. For EEG, CBraMod models patch dependencies through parallel spatial and temporal attention, while CSBrain combines cross-scale spatiotemporal modeling with structured sparse attention; both use masked reconstruction \citep{cbramod,csbrain}. LaBraM connects signal recovery to discrete prediction in two stages: Fourier amplitude and phase reconstruction first trains a vector-quantized tokenizer, which is then fixed to provide neural-code targets for masked prediction \citep{labram}. Reconstruction-derived information is passed through discrete targets, while reconstruction loss no longer directly optimizes the second-stage encoder. Re-JEPA also recovers amplitude and phase, but does so from continuous contextual predictions, allowing recovery gradients to flow through the predictor into the online encoder.

\subsection{JEPA-Based Self-Supervised Pretraining}
JEPA defines self-supervised objectives by predicting continuous target-region representations from visible context. I-JEPA applies this paradigm to images, and V-JEPA extends it to video, learning transferable features without pixel reconstruction \citep{ijepa,vjepa}. For physiological signals, EEG2Rep preserves contiguous subsequences to construct informative contexts and predict masked latent representations; EEG-JEPA combines structured electrode--time masks with multilevel target representations \citep{eeg2rep,eegjepa}. EMG-JEPA explores representation learning from sEMG without kinematic labels and transfer to cross-user hand pose estimation \citep{emgjepa}. Related methods also couple representation-level supervision with signal recovery. EEGPT jointly optimizes momentum-target alignment and masked reconstruction \citep{eegpt}. QualityFM combines self-distillation between quality-divergent ECG and PPG segments with amplitude and phase reconstruction from student representations \citep{guo2025qualityfm}. STST-JEPA predicts shallow representations from an EMA tokenizer and decodes masked predictions into signal segments \citep{ststjepa}. Re-JEPA connects \emph{contextual prediction in representation space} to \emph{attribute recovery in signal space} for continuous EMG tracking. Multilevel predictions are fused and restored to a complete token grid before amplitude and phase decoding. Objective comparisons, frozen readouts, and fine-tuning evaluate how these joint constraints support readability and generalization.

\section{Method}
\label{sec:method}
\begin{figure*}[!t]
\centering
\includegraphics[width=.96\textwidth]{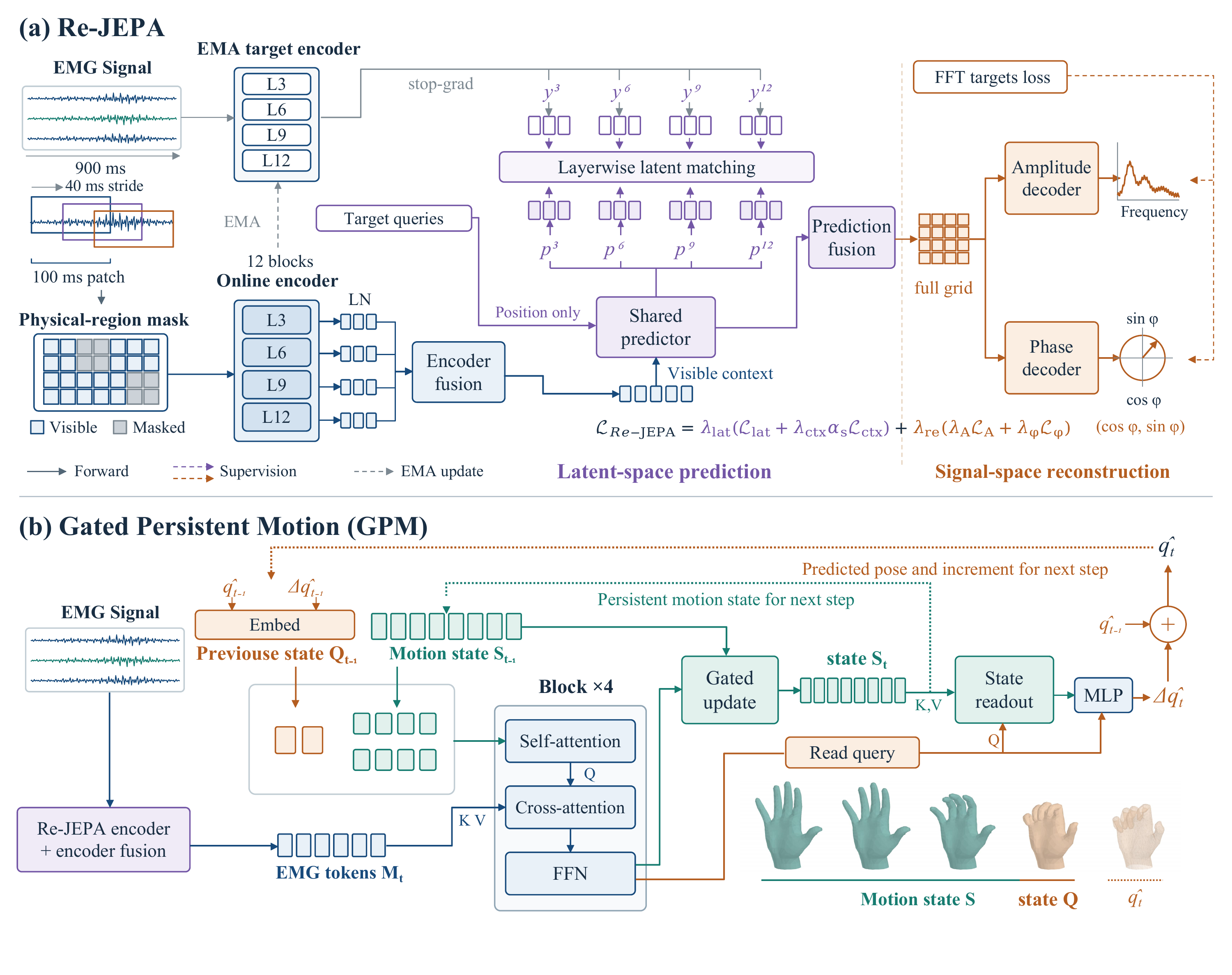}
\caption{Joint EMG pretraining and Gated Persistent Motion decoding. (a) Overlapping patches and physical-region masking feed an online encoder. Normalized L3/L6/L9/L12 features are fused and passed to a shared predictor with multilevel outputs, which match stop-gradient EMA targets. A separate fusion module restores the predicted token grid for independent amplitude and phase decoders. Solid lines indicate forward computation, colored dashed lines supervision, and gray dashed lines EMA updates. (b) Previous pose and increment tokens interact with persistent state tokens to read EMG memory, update the latent state through a gate, and predict an accumulated pose increment. Loops carry states and predictions to the next step. Only the encoder, level norms, and encoder fusion transfer from pretraining; GPM is randomly initialized. The hand sequence illustrates motion history and a prediction target, not measured prediction accuracy.}
\label{fig:framework}
\end{figure*}

\paragraph{Framework overview.}
As shown in \cref{fig:framework}, Re-JEPA predicts multilevel exponential-moving-average (EMA) target representations from masked EMG, fuses the predictions into a complete token grid, and decodes the masked signal's amplitude and phase attributes. Reconstruction gradients pass through the predictor to the online encoder, jointly constraining contextual prediction in representation space and signal recovery in signal space. For downstream tracking, the pretrained online encoder, level norms, and encoder fusion provide EMG tokens to GPM, which recurrently updates motion-state tokens and predicts pose increments.

The pretraining objective is
\begin{equation}
\mathcal L_{\rm pre}
=\lambda_{\rm lat}\mathcal L_{\rm lat}
 +\lambda_A\mathcal L_A+\lambda_\phi\mathcal L_\phi
 +\mathcal L_{\rm aux},
\label{eq:pre}
\end{equation}
where the first three terms supervise masked latent prediction, amplitude recovery, and phase recovery, respectively. $\mathcal L_{\rm aux}$ collects context consistency and variance--covariance regularization, detailed in \cref{app:aux}.

Our theoretical analysis shows that reconstruction rules out conditional complete collapse of the representation when its risk falls below the derived bound. If the decoder accesses sample content only through the predicted representation $Z$, with no signal bypass, we obtain the sufficient condition
\begin{equation}
\begin{aligned}
\mathcal R_A<\mathcal B_A\quad\Longrightarrow\quad
&Z\text{ is not conditionally}\\
&\text{completely collapsed}.
\end{aligned}
\label{eq:collapse_main}
\end{equation}
Here, $\mathcal R_A$ is the expected amplitude reconstruction loss and $\mathcal B_A$ is a pairwise risk lower bound constructed from independent targets under the same position and mask conditions. Conditional complete collapse means that representations no longer vary with sample content given these conditions. Definitions and the proof appear in \cref{app:collapse} (\cref{eq:d10}).

\subsection{EMG Tokenization and Physical-Region Masking}
Given an EMG window $X\in\R^{C\times L}$, we extract channel-wise patches of 100 ms with a 40 ms stride, yielding $16\times21=336$ tokens for a 900 ms, 16-channel input. Temporal convolutions map patches to $d=200$ embeddings with channel and time encodings. We denote the complete patch set by $\mathcal X$ and its indices by $\mathcal I$.

Masks are sampled as rectangles in channel--sample coordinates. Every patch intersecting their union $R$ is masked in full, defining target positions $\mathcal M$ and visible positions $\mathcal V=\mathcal I\setminus\mathcal M$. Target patches are zeroed before tokenization, and their embeddings are replaced by learned mask tokens. The online encoder processes the resulting full positional grid $\widetilde{\mathcal X}$, whereas the predictor reads only encoded visible positions. Visible patches therefore contain no raw samples inside $R$; for target patches crossing the boundary of $R$, samples outside $R$ may remain visible in overlapping unmasked patches. Mask sampling details appear in \cref{app:architecture}.

\subsection{Multilevel Joint-Embedding Predictive}
The online and EMA target encoders are 12-block Transformers with outputs at depths $\mathcal D=\{3,6,9,12\}$, each equipped with an independent learnable LayerNorm. The target encoder processes the complete input to produce normalized, stop-gradient targets $y_i^\ell$, incorporating full-window context. Its encoder parameters and level norms are updated by EMA.

At each visible position, normalized online features $h_i^\ell$ are concatenated across depths and fused:
\begin{equation}
f_i=F_{\rm enc}([h_i^\ell]_{\ell\in\mathcal D}),\qquad i\in\mathcal V.
\end{equation}
The fusion MLP maps $4d$ features to $d$. A shared four-block predictor processes these context tokens and position-encoded masked queries, which contain no target signal. Four output heads produce depth-specific predictions $p_i^\ell$ for both visible and masked positions. Masked predictions are matched to their corresponding target depths:
\begin{equation}
\mathcal L_{\rm lat}=\E_X\!\left[
\frac{\sum_{\ell\in\mathcal D}\sum_{i\in\mathcal M}
\rho(p_i^\ell,y_i^\ell)}{|\mathcal D||\mathcal M|}\right],
\end{equation}
where $\rho$ is feature-averaged Smooth L1. Target positions and depths are averaged within each example, followed by equal averaging across examples. Visible positions additionally receive context-consistency supervision (\cref{app:aux}). Normalization and EMA details appear in \cref{app:architecture}.

\subsection{Amplitude and Phase Reconstruction}
To constrain signal recoverability after contextual prediction, a separate fusion MLP combines the four predicted levels and restores the complete channel--time grid:
\begin{equation}
\begin{aligned}
r_i&=F_{\rm rec}([p_i^\ell]_{\ell\in\mathcal D}),\qquad i\in\mathcal I,\\
\widehat a&=D_A(\{r_i\}),\qquad
\widehat v=D_\phi(\{r_i\}).
\end{aligned}
\end{equation}
The amplitude and phase decoders each contain three Transformer blocks and an MLP head, with independent parameters. $F_{\rm rec}$ is separate from $F_{\rm enc}$. All visible and masked grid positions contain predictor outputs, without inserting clean target features.The reconstruction decoder processes this complete predicted grid rather than individual masked positions in isolation.

Targets are Fourier log amplitudes $a_{i,f}=\log(A_{i,f}+\epsilon)$ and unit-circle phases $u_{i,f}=(\cos\phi_{i,f},\sin\phi_{i,f})$ in the 20--850 Hz band. Amplitude recovery uses Smooth L1, while phase recovery uses an amplitude-weighted circular loss:
\begin{equation}
\begin{aligned}
\mathcal L_A&=\E_X\!\left[
\frac{\sum_{i\in\mathcal M,f\in\mathcal F}
\rho(\widehat a_{i,f},a_{i,f})}{|\mathcal M||\mathcal F|}\right],\\
\mathcal L_\phi&=\E_X\!\left[
\frac{\sum_{i\in\mathcal M,f\in\mathcal F}w_{i,f}
(1-\widehat u_{i,f}^{\mathsf T}u_{i,f})}
{\max(\sum_{i\in\mathcal M,f\in\mathcal F}w_{i,f},\epsilon)}\right].
\end{aligned}
\end{equation}
Here, $\mathcal F$ denotes the retained frequency bins, $\widehat u$ is the normalized phase prediction, and $w$ downweights low-amplitude frequencies using clipped relative amplitudes. Both losses supervise masked positions and average examples equally. Their gradients propagate through prediction fusion, the predictor, and encoder fusion into the online encoder. Spectral preprocessing and weight definitions appear in \cref{app:spectral}.

\subsection{Gated Persistent Motion Decoding}
\label{sec:gpm}
GPM maintains motion history in recurrent state tokens and conditions EMG interpretation on the previous predicted pose and increment (\cref{fig:framework}b). The pretrained online encoder, level norms, and $F_{\rm enc}$ provide downstream features. We select the last six temporal positions per channel to form EMG memory $M_k\in\R^{96\times d_g}$. These tokens cover 300 ms of raw signal while incorporating context from the complete encoded window.

At step $k$, the previous state $S_{k-1}\in\R^{K\times d_g}$ is concatenated with two tokens embedding the previous pose and increment:
\begin{equation}
\begin{aligned}
Q_k^{(0)}&=\LN\!\left([S_{k-1};e_q(\widehat q_{k-1});
e_\Delta(\Delta\widehat q_{k-1})]+E_{\rm role}\right),\\
Q_k&=U(Q_k^{(0)},M_k).
\end{aligned}
\end{equation}
Here, semicolons denote token concatenation and $E_{\rm role}$ identifies token roles. The updater $U$ contains four Transformer blocks, each applying query self-attention, cross-attention to $M_k$, and a feed-forward network. These operations combine latent motion history, explicit motion conditions, and current EMG observations. We use $K=8$ state tokens with $d_g=256$.

The first $K$ updated queries form a candidate state. An elementwise gate controls its incorporation into the persistent memory:
\begin{equation}
\begin{aligned}
\widetilde S_k&=\LN(Q_{k,1:K}),\\
G_k&=\sigma(W_g[S_{k-1},\widetilde S_k]+b_g),\\
S_k&=\LN\!\left(S_{k-1}+G_k\odot(\widetilde S_k-S_{k-1})\right).
\end{aligned}
\end{equation}
Commas denote feature concatenation. The updated state supports both the current prediction and subsequent recurrent steps.

The two updated motion-condition tokens jointly form a readout query, which attends to $S_k$ to predict the next pose increment:
\begin{equation}
\begin{aligned}
a_k&=e_r([Q_{k,K+1},Q_{k,K+2}]),\\
b_k&=\LN(\operatorname{Attn}(a_k,S_k,S_k)),\\
\Delta\widehat q_k&=\eta\,g_{\rm out}([b_k,a_k]),\\
\widehat q_k&=\widehat q_{k-1}+\Delta\widehat q_k.
\end{aligned}
\end{equation}
The two-layer MLP $g_{\rm out}$ outputs 20 joint-angle increments with scale $\eta=0.01$. Each rollout starts from a learned $S_0$, the provided initial pose, and a zero increment. Subsequent steps use predicted poses and increments without ground-truth substitution. State tokens persist across sliding windows within a rollout and reset between independent rollouts.

Training backpropagates through the complete recurrence using $\mathcal L_{\rm track}=\operatorname{MAE}_{\rm angle}+0.01E_{\rm fingertip}$, with angles in radians and fingertip distances in millimeters. Native 25 Hz encoder features are interpolated to 50 Hz before recurrence, and predictions are aligned to valid labels. Initialization and temporal alignment details appear in \cref{app:gpm}; frozen-readout and fine-tuning protocols are described in \cref{sec:experiments}.

\section{Experiments}
\label{sec:experiments}
\begin{table*}[t]
\centering
\caption{Tracking test results with the initial pose provided. Values are user-level means $\pm$ standard deviations; lower is better.}
\label{tab:main-tracking}
\footnotesize
\setlength{\tabcolsep}{1.3pt}
\renewcommand{\arraystretch}{1.15}
\begin{tabular}{@{}lccccccc@{}}
\toprule
\multirow{2}{*}{Model} & \multirow{2}{*}{Training} & \multicolumn{2}{c}{Cross-user} & \multicolumn{2}{c}{Cross-stage} & \multicolumn{2}{c}{Cross-user\,+\,stage} \\
\cmidrule(lr){3-4}\cmidrule(lr){5-6}\cmidrule(l){7-8}
& & MAE ($^\circ$) & Landmark (mm) & MAE ($^\circ$) & Landmark (mm) & MAE ($^\circ$) & Landmark (mm) \\
\midrule
vemg2pose & Supervised & $7.70 \pm 1.00$ & $10.30 \pm 1.50$ & $11.20 \pm 1.40$ & $15.20 \pm 1.90$ & $11.00 \pm 1.00$ & $15.40 \pm 1.40$ \\
NHN & Supervised & $7.30 \pm 0.88$ & $9.55 \pm 1.41$ & $10.12 \pm 1.31$ & $13.59 \pm 1.69$ & $10.16 \pm 0.99$ & $14.03 \pm 1.33$ \\
EMG-GPT$^{\dagger}$ & Fine-tuned & $8.89 \pm 0.93$ & $11.59 \pm 1.24$ & $12.42 \pm 1.43$ & $16.88 \pm 1.80$ & $12.18 \pm 1.25$ & $16.93 \pm 1.36$ \\
LaBraM (L12) + GPM & Frozen & $7.87 \pm 1.01$ & $10.60 \pm 1.59$ & $11.77 \pm 1.45$ & $16.23 \pm 1.83$ & $11.44 \pm 1.08$ & $16.24 \pm 1.38$ \\
LaBraM + LSTM & Fine-tuned & $7.48 \pm 0.95$ & $9.79 \pm 1.51$ & $10.44 \pm 1.34$ & $13.90 \pm 1.65$ & $10.44 \pm 1.06$ & $14.42 \pm 1.44$ \\
LaBraM (L12) + GPM & Fine-tuned & $7.27 \pm 0.87$ & $9.33 \pm 1.38$ & $10.01 \pm 1.31$ & $13.14 \pm 1.62$ & $10.00 \pm 0.94$ & $13.51 \pm 1.24$ \\
\midrule[1.2pt]
Re-JEPA (Fusion) + GPM & Frozen & $7.23 \pm 0.83$ & $9.23 \pm 1.35$ & $9.88 \pm 1.28$ & $12.92 \pm 1.61$ & $10.07 \pm 0.88$ & $13.57 \pm 1.15$ \\
Re-JEPA (L9) + GPM & Frozen & $7.11 \pm 0.84$ & $8.99 \pm 1.36$ & $9.54 \pm 1.26$ & $12.37 \pm 1.58$ & $9.84 \pm 0.90$ & $13.16 \pm 1.18$ \\
Re-JEPA (L12) + GPM & Frozen & $7.14 \pm 0.85$ & $9.10 \pm 1.33$ & $9.70 \pm 1.27$ & $12.65 \pm 1.54$ & $9.87 \pm 0.91$ & $13.23 \pm 1.21$ \\
\midrule
Re-JEPA (Fusion) + LSTM & Fine-tuned & $7.30 \pm 0.84$ & $9.37 \pm 1.33$ & $9.60 \pm 1.28$ & $12.42 \pm 1.51$ & $9.95 \pm 0.93$ & $13.39 \pm 1.17$ \\
Re-JEPA w/o VCReg + GPM & Fine-tuned & $6.94 \pm 0.84$ & $8.71 \pm 1.27$ & $\mathbf{9.02} \pm 1.23$ & $\mathbf{11.61} \pm 1.52$ & $9.47 \pm 0.87$ & $12.56 \pm 1.12$ \\
\textbf{Re-JEPA (Fusion) + GPM} & Fine-tuned & $\mathbf{6.91} \pm 0.83$ & $\mathbf{8.69} \pm 1.27$ & $9.04 \pm 1.16$ & $11.64 \pm 1.41$ & $\mathbf{9.36} \pm 0.90$ & $\mathbf{12.47} \pm 1.12$ \\
\bottomrule
\end{tabular}
\par\smallskip
\begin{minipage}{\textwidth}\footnotesize
Published results: emg2pose Table~5 \citep{emg2pose}, NHN Table~1(b) \citep{nhn}, and EMG-GPT Appendix~E Table~4 \citep{emggpt}.
$^{\dagger}$EMG-GPT uses a different scoring grid and is shown for reference only.
NHN averages five seeds per user before computing user-level statistics.
For our evaluations, SD uses the $n-1$ denominator, with $n=20/158/20$ users across the three splits.
Frozen denotes training only GPM while keeping the pretrained representation path fixed. For GPM readout, LaBraM uses 48 L12 tokens per decoding step, whereas Re-JEPA uses 96 tokens from Fusion unless L9 or L12 is explicitly specified. Fusion combines L3, L6, L9, and L12. Both token sets cover the final 300 ms of raw signal but are extracted from different token grids. Fine-tuned LaBraM models update the pretrained representation path from epoch zero; Re-JEPA+GPM first trains GPM for four frozen-encoder epochs before full fine-tuning. Re-JEPA+LSTM mean-pools the same 96-token Fusion input and fine-tunes from epoch zero. Comparisons between LaBraM and Re-JEPA therefore evaluate complete systems with different tokenization, readout, and training configurations, rather than isolate the pretraining objective.
The w/o VCReg variant removes pretraining variance/covariance penalties, retains context consistency, and uses Fusion+GPM96.
Bold values mark the lowest means per column, excluding EMG-GPT, not statistical significance.
\end{minipage}
\end{table*}

Full fine-tuning is our primary evaluation of cross-user, cross-stage, and joint generalization. The J versus J+R comparison tests recovery constraints after contextual prediction; frozen readouts, decoder comparisons, and representation interventions provide complementary evidence. Results and interpretation appear in \cref{sec:results,sec:discussion}, with complete metrics and implementation details in the appendices.

\subsection{Pretraining Setup}
We pretrain on EMG from the official emg2pose training split without using kinematic labels. The full dataset contains approximately 370 hours of recordings from 193 users across 29 motion stages \citep{emg2pose}. Validation and test recordings are excluded from parameter updates. Re-JEPA transfers the final pretrained online encoder, level norms, and encoder fusion to downstream tracking. Complete training configurations are provided in \cref{app:optimization}.

\subsection{Data Preprocessing}
We retain the native 16-channel EMG at 2 kHz without additional per-window amplitude normalization. Patch extraction and physical-region masking follow Section~3.1. Training uses random circular channel shifts, whereas validation and testing omit this augmentation. Spectral processing constructs reconstruction targets without altering encoder inputs.

For downstream tracking, Re-JEPA and LaBraM read 96 and 48 tokens per step, respectively. Both token sets cover the final 300 ms of raw signal while incorporating full-window context. Supervision and evaluation use time-aligned valid labels only. Window sampling, augmentation parameters, padding, and temporal alignment are detailed in \cref{app:architecture}.

\subsection{Evaluation Protocol}
\paragraph{Tracking.}
We use the official cross-user, cross-stage, and joint splits, providing the ground-truth pose only at the beginning of each evaluation segment \citep{emg2pose}. In our evaluations, tracking proceeds in independent 5 s segments. Each segment is initialized with its ground-truth pose, after which the decoder predicts in closed loop using its own pose estimates, without further ground-truth pose inputs. Recurrent states are reset between segments, so motion memory is not carried across segment boundaries. Angular MAE (degrees) and landmark error (millimeters) are computed per user and then averaged with equal user weight. Complete metrics, including fingertip error, appear in \cref{app:extended}.

\paragraph{Fine-tuning and frozen readouts.}
Re-JEPA+GPM freezes the pretrained representation path for four epochs before joint fine-tuning; Re-JEPA+LSTM and LaBraM fine-tune from the start. Frozen-readout experiments keep the representation path fixed throughout training and optimize only GPM. Checkpoints are selected by validation loss. Training configurations are detailed in \cref{app:optimization,app:decoder}.

\paragraph{Comparisons.}
We distinguish published results for vemg2pose, NHN, and EMG-GPT from results obtained by training and evaluating an EMG-adapted LaBraM baseline in this study. EMG-GPT is included for reference because its scoring and temporal alignment protocol differs. \cref{tab:main-tracking} compares complete tracking systems, with model-specific tokenization, readout layers, and downstream training schedules documented in its notes. The J versus J+R comparison is reported separately under the same Fusion+GPM96 downstream interface and test aggregation. We also evaluate LSTM and GPM configurations for both LaBraM and Re-JEPA.

\paragraph{Representation analysis.}
Fixed-encoder MLP and lightweight Transformer readouts use recording-disjoint subsets of the training partition. Dynamic readouts receive the current true pose $q_t$ and are evaluated separately from closed-loop tracking. Context and query perturbations measure absolute and normalized representation changes without a pose decoder. \Cref{fig:analysis} summarizes these analyses; complete protocols appear in \cref{app:readout,app:analysis}.

\section{Results}
\label{sec:results}
\subsection{Tracking Performance and Generalization}
Fine-tuned Re-JEPA+GPM achieves angular MAEs of 6.91$^\circ$, 9.04$^\circ$, and 9.36$^\circ$ on cross-user, cross-stage, and joint splits (\cref{tab:main-tracking}). These are 10.3\%, 19.3\%, and 14.9\% below published vemg2pose means and 4.9\%, 9.7\%, and 6.5\% below LaBraM+GPM, with improvements also over NHN. Landmark errors follow the same trend. The LaBraM comparison evaluates complete systems with different token grids and fine-tuning schedules; checkpoint details and full metrics appear in \cref{app:extended}.

Removing variance and covariance penalties changes angular MAE by approximately $+0.03^\circ$, $-0.02^\circ$, and $+0.11^\circ$, retaining similar tracking performance. \Cref{app:collapse} relates this ablation to the collapse analysis.

\subsection{Effect of Reconstruction after Latent Prediction}
Under the same Fusion+GPM96 interface and tracking protocol, J+R lowers angular MAE over J by 4.6\%, 8.4\%, and 7.0\% (\cref{tab:objectives}). Landmark errors also decrease across all splits, supporting signal recovery after latent prediction, with larger relative gains on unseen motion stages.

\begin{table*}[t]
\centering
\caption{Pretraining objectives under the same Fusion+GPM96 downstream interface. Values are user-level means $\pm$ standard deviations; lower is better.}
\label{tab:objectives}
\small
\begin{tabular}{@{}lcccccc@{}}
\toprule
&\multicolumn{2}{c}{Cross-user}&\multicolumn{2}{c}{Cross-stage}&\multicolumn{2}{c}{Cross-user+stage}\\
\cmidrule(lr){2-3}\cmidrule(lr){4-5}\cmidrule(l){6-7}
Objective&MAE ($^\circ$)&Landmark (mm)&MAE ($^\circ$)&Landmark (mm)&MAE ($^\circ$)&Landmark (mm)\\
\midrule
J&$7.24\pm0.85$&$9.27\pm1.34$&$9.87\pm1.34$&$12.88\pm1.65$&$10.06\pm0.95$&$13.52\pm1.22$\\
\textbf{J+R}&$\mathbf{6.91}\pm0.83$&$\mathbf{8.69}\pm1.27$&$\mathbf{9.04}\pm1.16$&$\mathbf{11.64}\pm1.41$&$\mathbf{9.36}\pm0.90$&$\mathbf{12.47}\pm1.12$\\
\bottomrule
\end{tabular}
\end{table*}

\subsection{Frozen-Encoder Tracking}
Training only GPM on fixed Re-JEPA Fusion features lowers angular MAE by 8.1\%, 16.1\%, and 12.0\% relative to frozen LaBraM (\cref{tab:main-tracking}). L9 outperforms both L12 and Fusion, showing that readability does not increase monotonically with depth or fusion. The main configuration retains its predefined Fusion interface. Complete layer-wise results appear in \cref{app:frozen}.

\subsection{GPM Decoder Evaluation}
GPM lowers angular MAE relative to LSTM by 2.9\%, 4.1\%, and 4.2\% on LaBraM and by 5.4\%, 5.8\%, and 6.0\% on Re-JEPA; landmark errors also improve (\cref{tab:main-tracking}). The Re-JEPA heads share E399 initialization, 96 Fusion tokens, and the test protocol. LSTM mean-pools tokens and fine-tunes directly, whereas GPM preserves tokens for state-conditioned reading and starts with four frozen epochs. These results compare readout-and-training configurations rather than isolate individual GPM components (\cref{app:decoder}).

\subsection{Representation Readout and Context Dependence}
\Cref{fig:analysis} compares fixed-encoder motion-stage and 40 ms pose-increment readouts with $K=1/4$ histories. Dynamic readouts receive the true current pose and use recording-disjoint evaluation, separate from closed-loop tracking (\cref{app:readout}). For the $K=4$ Transformer, Re-JEPA, LaBraM, and latent-only JEPA achieve action accuracies of 37.18\%, 26.11\%, and 27.68\%, and vector-MAE gains over the training-label-mean predictor of 8.82\%, 1.43\%, and 2.78\%. MLP vector readouts generally match or underperform this baseline; head comparisons also differ in training budget.

With the query signal and entry embedding fixed, paired donor perturbations to 12 neighboring patches yield higher Re-JEPA relative RMS responses at L3/L6/L9/L12 under both 25\% and 100\% mixing. Responses are normalized by each model's natural representation differences. At L12 and 100\% mixing, Re-JEPA reaches 95.05\% versus LaBraM's 54.59\% (\cref{app:perturbation}). This identifies stronger conditional context dependence; its causal contribution to tracking gains remains unresolved.

\section{Discussion}
\label{sec:discussion}
\begin{figure}[!t]
\centering
\includegraphics[width=\columnwidth]{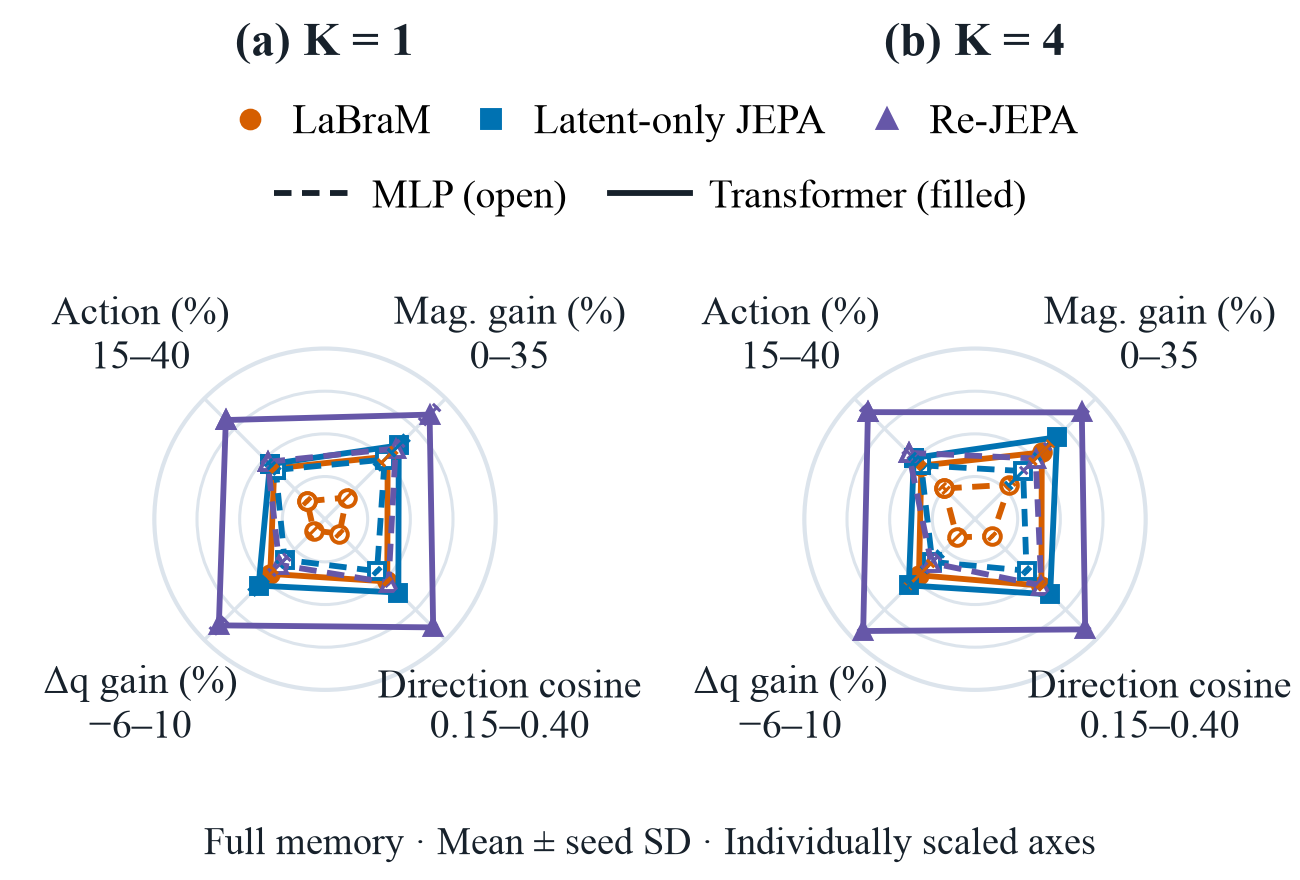}
\includegraphics[width=\columnwidth]{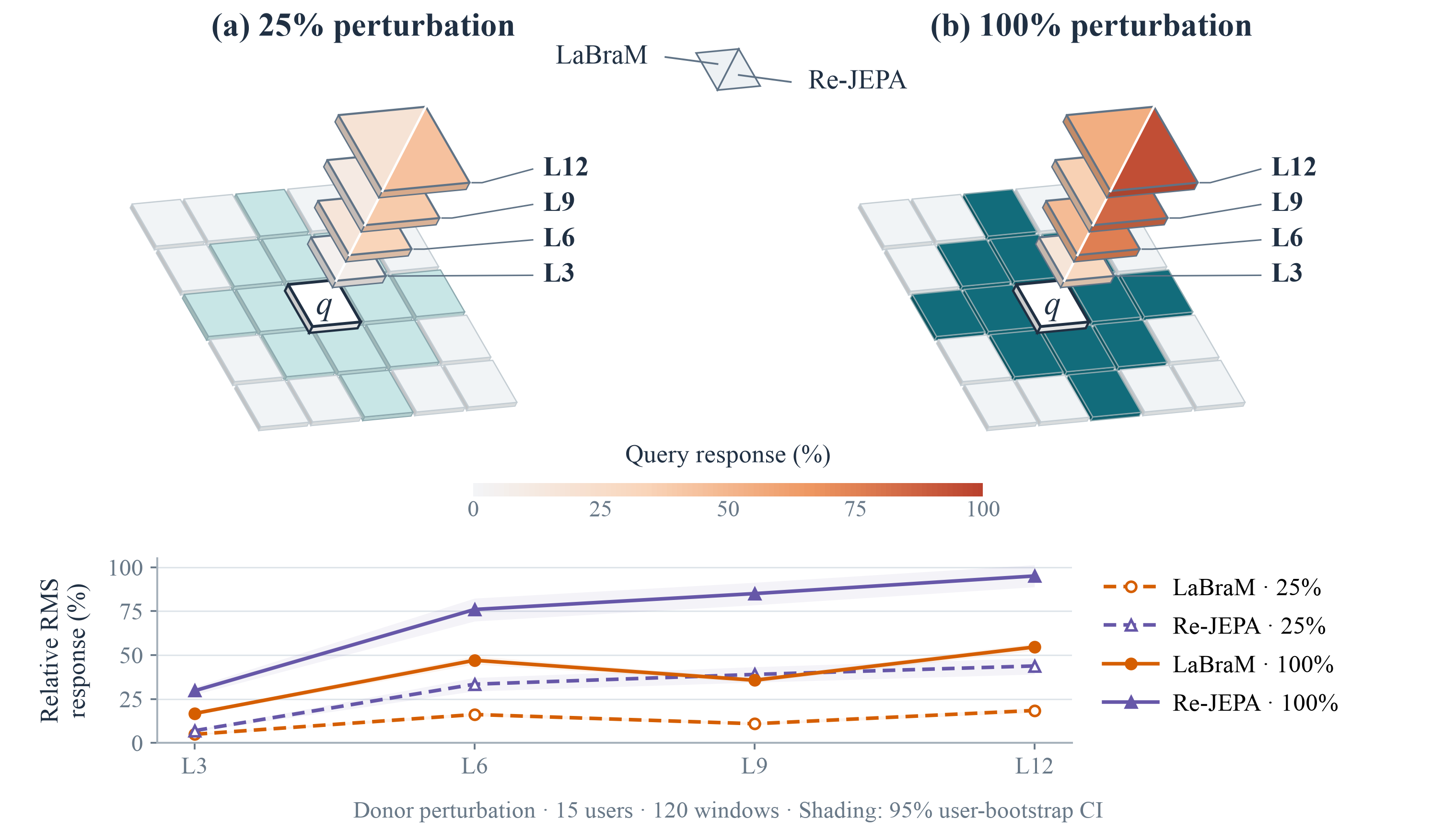}
\caption{Motion readout and context dependence with fixed representations. Top: $K=1/4$ readouts; colors identify models, dashed/open markers denote MLPs, and solid/filled markers denote Transformers. Error bars show readout-seed SD. Axes are independently scaled with identical per-metric ranges in both radars; area is not a composite score. Bottom: paired interventions on 12 neighboring patches at 25\%/100\% donor mixing (not coverage). Each layer's upper-left/lower-right triangle denotes LaBraM/Re-JEPA under a shared response scale. Curves aggregate 120 windows from 15 users; bands are 95\% user-bootstrap intervals.}
\label{fig:analysis}
\end{figure}

We study how signal recoverability complements continuous contextual prediction for EMG tracking. Joint pretraining improves fine-tuned tracking and fixed-encoder motion readouts over latent-only pretraining (\cref{tab:objectives,fig:analysis}), locating an advantage before downstream encoder adaptation. The tracking comparison uses the same Fusion+GPM96 interface and test aggregation, but equivalence in pretraining data exposure, auxiliary loss settings, and training budgets has not yet been fully verified. These results demonstrate an advantage for the evaluated joint configuration, without isolating reconstruction as the sole source of the gains.

Context interventions further reveal stronger query responses to neighboring inputs relative to each model's natural representation differences. This is consistent with learning cross-patch prediction relationships. Importantly, L9 achieves the lowest frozen-encoder tracking error among the evaluated L9, L12, and Fusion readouts, although its normalized context response is not the largest. L12 responds more strongly to the tested contextual perturbations but performs worse in frozen GPM tracking (\cref{tab:main-tracking,fig:analysis}). This contrast distinguishes context sensitivity from task readability: stronger responses to neighboring inputs do not necessarily make a representation better suited to tracking. Context sensitivity characterizes how representations depend on surrounding signals, whereas tracking performance also depends on whether the resulting information can be effectively read out for motion estimation.

GPM provides a task-specific way to read these representations. Our LSTM baseline mean-pools EMG tokens before recurrent decoding, whereas GPM preserves them for motion-conditioned attention. Persistent state tokens carry learned motion memory, explicit pose and increment tokens specify the current estimate, and a gate controls how new EMG evidence updates that memory before the next pose increment is predicted. The distinction is therefore state-conditioned selection of token-level evidence, rather than memory alone. GPM configurations reduce angular MAE by 5.4\%, 5.8\%, and 6.0\% on Re-JEPA, with consistent improvements on LaBraM (\cref{tab:main-tracking}), supporting the value of this readout design for continuous tracking.

Reconstruction also supplies a constraint on complete collapse through targets determined by observations rather than the evolving momentum encoder. Under the no-signal-bypass assumption, \cref{app:collapse} derives a reconstruction-risk floor for representations that no longer vary with sample content given position and mask information. Risk below this floor excludes conditional complete collapse. Empirically, removing variance and covariance penalties changes angular MAE by $+0.032^\circ$, $-0.025^\circ$, and $+0.108^\circ$, all below 1.2\% in relative magnitude. These small, mixed changes indicate limited additional tracking benefit from explicit regularization in the current joint configuration. The theory and ablation provide complementary evidence for a recoverability constraint and reduced reliance on these penalties.

Current results establish benefits for EMG tracking but do not identify which fine-grained signal features reconstruction preserves, which dominant factors contextual prediction retains, or how these factors contribute to task performance. Context sensitivity alone cannot distinguish task-relevant dependencies from predictable nuisance variation. The balance between suppressing task-irrelevant detail and preserving useful information across representation and signal spaces therefore remains unresolved. Its appropriate form may also depend on the signal's physiological origin, noise structure, and downstream task; whether other physiological signals benefit from the amplitude-and-phase constraints used here is untested. The collapse result likewise requires empirical estimation of the conditional reconstruction-risk floor, while the no-VCReg ablation retains other stabilizing mechanisms and does not isolate reconstruction's contribution.

Future work will extend the framework to other physiological and time-series signals while studying this information trade-off explicitly. Controlled interventions on temporal, spectral, and acquisition-related factors, combined with analyses of retained and recoverable information, could help distinguish useful signal detail from nuisance variation. These findings will guide models that adapt prediction targets and reconstruction constraints to signal characteristics and task information needs, including which attributes to recover and at what granularity. We will also examine how this balance affects generalization and empirically test the conditional risk criterion to clarify reconstruction's role in preventing complete collapse.

\FloatBarrier
\section*{Impact Statement}
This work studies self-supervised representations for continuous hand tracking from surface electromyography. Such representations may support non-visual interfaces and the collection of human motion demonstrations. Physiological recordings and inferred motion can expose personal information; their collection and use require appropriate consent and access controls. The reported tracking setting provides an initial pose and does not establish clinical validity or safety for controlling physical devices. Deployment should account for user variability, prediction failures, and the temporal alignment used by the decoder.
\bibliography{references-arxiv,references-other}
\bibliographystyle{icml2026}
\clearpage
\appendix
\numberwithin{equation}{section}
\onecolumn
\raggedbottom
\section{Auxiliary Objectives and Loss Reduction}
\label{app:aux}
\subsection{Context Consistency}
Visible-position matching is weighted by distance to the nearest target on the index grid:
\begin{equation}
\mathcal L_{\rm ctx}=\E_X\!\left[
\frac{1}{|\mathcal D||\mathcal V|}
\sum_{\ell\in\mathcal D}\sum_{i\in\mathcal V}
d(i,\mathcal M)^{-1/2}\rho(p_i^\ell,y_i^\ell)\right],
\end{equation}
\begin{equation}
d(i,\mathcal M)=\max\!\left(1,\min_{j\in\mathcal M}
\{|c_i-c_j|+|t_i-t_j|\}\right).
\end{equation}
This is Manhattan distance on channel--time indices, not physical electrode distance or circular shortest-channel distance. Weighted errors are divided by the number of visible positions; weights are not renormalized to sum to one. Masks are sampled independently for each example. Padding introduced when packing examples contributes neither attention keys/values nor any loss statistic.

\subsection{Variance and Covariance Regularization}
For depth $\ell$, valid visible and target predictions on the current device form $P^\ell\in\R^{n_\ell\times d}$, truncated to the first 8192 rows. Define
\begin{equation}
\begin{aligned}
\mathcal L_{\rm var}&=\frac{1}{|\mathcal D|d}\sum_{\ell,j}
\max\!\left(0,\gamma-\sqrt{\operatorname{Var}(P^\ell_{\cdot j})+10^{-4}}\right),\\
C^\ell&=\frac{(P^\ell-\mathbf1\mu_\ell^{\mathsf T})^{\mathsf T}
(P^\ell-\mathbf1\mu_\ell^{\mathsf T})}{\max(1,n_\ell-1)},\\
\mathcal L_{\rm cov}&=\frac{1}{|\mathcal D|d}\sum_\ell\sum_{j\ne k}(C^\ell_{jk})^2.
\end{aligned}
\end{equation}
$\mu_\ell$ is the feature mean, $\operatorname{Var}$ uses the population denominator, and $\gamma=1$. Statistics are device-local, with no cross-device feature all-gather. The 8192-row cap is sequential truncation, not random subsampling.

\subsection{Weights and Aggregation}
The auxiliary term in \cref{eq:pre} is
\begin{equation}
\mathcal L_{\rm aux}
=\lambda_{\rm lat}\lambda_{\rm ctx}\alpha_s\mathcal L_{\rm ctx}
 +\beta_s(\lambda_{\rm var}\mathcal L_{\rm var}
 +\lambda_{\rm cov}\mathcal L_{\rm cov}).
\end{equation}
\begin{center}
\small
\begin{tabularx}{\textwidth}{@{}lcY@{}}
\toprule
Objective & Main weight & Reduction\\
\midrule
Target latent $\mathcal L_{\rm lat}$ & 1 & Equal layers; target-token mean within each example\\
Context $\mathcal L_{\rm ctx}$ & $0.5\alpha_s$ & Equal layers; visible-token mean of distance-weighted errors\\
Amplitude $\mathcal L_A$ & 1 & Target-token and frequency-bin mean within each example\\
Phase $\mathcal L_\phi$ & 1 & Weighted normalization across target tokens and frequencies within each example\\
Variance $\mathcal L_{\rm var}$ & $10^{-3}\beta_s$ & Device-local valid predictions; equal layers\\
Covariance $\mathcal L_{\rm cov}$ & $10^{-4}\beta_s$ & Device-local valid predictions; equal layers\\
\bottomrule
\end{tabularx}
\end{center}
Both $\alpha_s$ and $\beta_s$ increase linearly with optimization steps over the first ten epochs and remain at one thereafter. Latent, context, amplitude, and phase losses are first reduced per example and then averaged equally across examples; a single batch-wide token mean is not equivalent. Smooth L1 uses transition parameter 1. All objectives are optimized in the same forward/backward pass.

\section{Architecture and Reproducibility}
\label{app:architecture}
\subsection{Pretraining Architecture and Masking}
\begin{center}
\small
\begin{tabularx}{\textwidth}{@{}lY@{}}
\toprule
Component & Main configuration\\
\midrule
Input and grid & $16\times1800$ samples at 2 kHz; patch length 200, stride 80; $16\times21=336$ patch tokens\\
Online/target encoders & 12 LaBraM-style blocks, width 200, 10 heads, MLP ratio 4, query/key normalization\\
Multilevel outputs & L3/L6/L9/L12, each with an independent learnable LayerNorm\\
$F_{\rm enc}$ & Linear(800,400) $\to$ GELU $\to$ Linear(400,200) $\to$ LayerNorm\\
Predictor & Width 100; four blocks, four heads, FFN width 400; four Linear(100,200) heads\\
$F_{\rm rec}$ & LayerNorm(800) $\to$ Linear(800,400) $\to$ GELU $\to$ Linear(400,200) $\to$ LayerNorm\\
Amplitude/phase decoders & Three blocks each, width 200, 10 heads; independent parameters\\
Amplitude/phase heads & $200\to200\to84$ / $200\to200\to168$, with Tanh hidden activation\\
Position encoding & Learned channel/time embeddings in the encoder; fixed 2D sine/cosine encoding in the predictor\\
Input scaling & Online-input scale 1; no additional per-window amplitude normalization\\
\bottomrule
\end{tabularx}
\end{center}
The encoder retains a CLS token for attention but does not use it as a patch-prediction target or downstream memory token. Fusion operates independently at each token position, without averaging over time or channels.

Pretraining uses 900 ms windows sampled at a nominal interval of 900 ms, with start-position jitter.

Let $E_\theta$ and $E_{\bar\theta}$ denote the online and target encoders, with corresponding learnable level norms $N_\theta^\ell$ and $N_{\bar\theta}^\ell$. Their normalized features are
\begin{equation}
h_i^\ell=N_\theta^\ell(E_\theta^\ell(\widetilde{\mathcal X})_i),\qquad
y_i^\ell=\sg\!\left[\LN_0\!\left(N_{\bar\theta}^\ell(E_{\bar\theta}^\ell(\mathcal X)_i)\right)\right].
\end{equation}
$\LN_0$ normalizes over features without affine parameters, and $\sg$ stops gradients. The target encoder and its level norms are initialized from the online counterparts and updated at step $s$ by
\begin{equation}
\bar\theta\leftarrow m_s\bar\theta+(1-m_s)\theta,
\end{equation}
where both parameter sets include the corresponding level norms. Predictor and fusion parameters are not EMA-updated. Visible features and learned masked queries both receive two-dimensional channel--time encodings before the shared predictor. Its four independent output heads predict $p_i^\ell\in\R^d$ for all visible and masked positions within the same window.

For patch length $P$ and stride $S$, the patch extraction and index grid are
\begin{equation}
x_{c,t}=X_{c,tS:tS+P},\qquad
T=1+\left\lfloor\frac{L-P}{S}\right\rfloor,\qquad
\mathcal I=\{1,\ldots,C\}\times\{0,\ldots,T-1\}.
\label{eq:patch}
\end{equation}
The complete patch set is $\mathcal X=\{x_i:i\in\mathcal I\}$. For the union $R$ of sampled physical regions, the target set is $\mathcal M=\{i\in\mathcal I:\operatorname{supp}(x_i)\cap R\ne\varnothing\}$ and the visible set is $\mathcal V=\mathcal I\setminus\mathcal M$. Target patch values are zeroed before convolution and normalization, and the resulting embeddings are replaced by a learned mask token.

Each example independently samples six rectangles, each spanning four adjacent channels and 400 samples. Channel indices wrap around; temporal intervals do not. Overlapping rectangles retain their union rather than being trimmed to a token budget. If no visible or no target position remains, the entire rectangle set is resampled. Every patch intersecting the union is masked and supervised, including boundary patches. Raw masked area, target-token fraction, and boundary fraction are distinct quantities. The legacy mask-ratio configuration field does not determine this branch's effective masking rate.

\subsection{Spectral Targets}
\label{app:spectral}
With a periodic Hann window $w[n]$ of length 200, each original patch is independently demeaned before transformation:
\begin{equation}
Z_{i,f}=\operatorname{rFFT}_f\!\left((x_i-\operatorname{mean}(x_i))\odot w\right),
\qquad A_{i,f}=\frac{\kappa_f|Z_{i,f}|}{\sum_nw[n]}.
\end{equation}
$\kappa_f=1$ at DC and Nyquist and 2 elsewhere. The retained 20--850 Hz band has 84 bins at 10 Hz spacing; all retained bins use the single-sided doubling correction. Log amplitude is not further standardized by patch or window.

Phase targets are real/imaginary vectors obtained by dividing the complex spectrum by its magnitude, with denominator floor $\epsilon=10^{-6}$. Target and predicted vectors are stably normalized by their two-dimensional norm before the loss. Zero spectra yield zero target vectors and zero phase weights. The reliability weights are
\begin{equation}
w_{i,f}=\sg\!\left[\min\!\left(4,
\frac{A_{i,f}}{\max(\overline A_i,\epsilon)}\right)\right],\qquad
\overline A_i=\frac{1}{|\mathcal F|}\sum_{f\in\mathcal F}A_{i,f}.
\end{equation}
They divide amplitude by its patch's band mean, clip at 4, and are normalized over all target positions and bins within an example in the phase loss. This is neither an equal-weighted per-patch phase loss nor MSE on wrapped phase angles.

Both decoders use LaBraM-style Transformer blocks and read the complete predicted grid, while losses use target positions only. Reconstruction gradients do not enter the stop-gradient EMA branch. There is no extra time-domain waveform head; the implementation's \texttt{time\_decoder} denotes the phase decoder. Demeaning, windowing, and band selection mean that these targets do not imply lossless recovery of the original waveform.

\subsection{GPM Update and Temporal Alignment}
\label{app:gpm}
Downstream inference retains the online encoder, level norms, and $F_{\rm enc}$, but not the pretraining predictor, $F_{\rm rec}$, or reconstruction decoders. Selected token features are projected, augmented with learned channel and relative-time embeddings, and normalized to form $M_k$. Independent linear maps embed the previous 20-dimensional pose and increment as condition tokens. The shared learned initial state is reset at each rollout, with $\widehat q_0=q_0$ and $\Delta\widehat q_0=0$.

Re-JEPA selects the final six temporal tokens per channel (96 total), whereas LaBraM selects the final three (48 total). Both selected raw supports span the final 300 ms, with contextual information from the complete encoded window.

For input queries $Q$ and current memory $M_k$, each update block uses pre-normalization and residual connections \citep{attention}:
\begin{equation}
\begin{aligned}
Q'&=Q+\operatorname{MHA}_{\rm self}(\LN_1(Q),\LN_1(Q),\LN_1(Q)),\\
Q''&=Q'+\operatorname{MHA}_{\rm cross}(\LN_2(Q'),\LN_M(M_k),\LN_M(M_k)),\\
Q'''&=Q''+\operatorname{FFN}(\LN_3(Q'')).
\end{aligned}
\end{equation}
Four independently parameterized eight-head blocks update the queries sequentially; memory is not recurrently updated between these blocks. FFN width is $4d_g$ and dropout is zero. Eight state tokens plus pose and increment conditions give ten queries, attending to 96 memory tokens. Only state tokens persist as latent memory; the two motion conditions are rebuilt from previous predictions. Each output increment corresponds to one recurrent update, rather than angular velocity normalized by seconds.

The learned initial state uses truncated-normal initialization with standard deviation 0.02, as do channel, relative-time, and role embeddings. Gate weights start at zero and biases at $-1$, giving an initial sigmoid value of approximately 0.269. The following LayerNorm means that the complete update is not simply retention of a fixed fraction of the raw state. All GPM parameters are randomly initialized, with no downstream checkpoint loaded.

Each encoded window contains 1791 raw samples and nine left-padding zeros, matching the pretrained 1800-sample grid; windows advance by 80 samples. A training segment contains 11,790 samples: 1790 left-context samples and 5 s of valid motion. Its 125 native feature frames are linearly interpolated along the window dimension to 250 GPM steps with \texttt{align\_corners=True}. This interpolation can mix neighboring windows, so zero encoder right context alone does not ensure causality of the full decoding path. Token positions are neither interpolated nor mean-pooled.

Each independent rollout resets the latent state and initial increment. The initial pose comes from the valid segment start; later ground-truth poses are not teacher-forced. State, pose, and increment recurrences are not detached during training: gradients traverse the complete rollout. Predictions are interpolated to the label grid before valid-label loss computation. Fingertip error uses the same hand kinematic model, sampled every 40 label points. Pose conditions, rollout segmentation, and aggregation must be kept consistent across evaluations.

\subsection{Optimization Settings}
\label{app:optimization}
The EMA coefficient increases linearly from 0.996 to 1 during pretraining. Main-model pretraining uses 48 NVIDIA A800 GPUs.
\begin{center}
\small
\begin{tabularx}{\textwidth}{@{}lYY@{}}
\toprule
Setting & Unlabeled pretraining & Main downstream training\\
\midrule
Initialization & Random online branch; target copied from online modules & E399 online encoder, level norms, $F_{\rm enc}$; random GPM\\
Global batch & 4608 ($48\times96$) & 192 ($48\times4$)\\
Optimizer & AdamW, $\beta=(0.9,0.98)$, $\epsilon=10^{-8}$, weight decay 0.05 & AdamW, $\epsilon=10^{-8}$, weight decay 0.05\\
Learning rate & 0.003; 50-epoch warmup; cosine to $10^{-5}$ & 0.000612372435696; 5-epoch warmup; cosine to $10^{-6}$\\
Budget & 400 epochs; gradient-norm clip 1 & Up to 50 epochs; frozen for the first four\\
Layer/group rates & Not applicable & Layer decay 0.8; encoder multiplier 1; fusion multiplier 0.25\\
Hierarchy dropout & None & 0 while frozen; 0.1 during full fine-tuning\\
Precision/seed & bf16 mixed / 42 & bf16 mixed / 42\\
Selection/stopping & Final E399 & Validation loss; minimum improvement 0.0005, patience 5; top three plus last\\
\bottomrule
\end{tabularx}
\end{center}
The frozen phase fixes the encoder, level norms, and $F_{\rm enc}$ and optimizes only GPM. Frozen-readout comparisons train for at most 20 epochs and never enter an unfreezing phase. Training augmentation circularly shifts channels by $-1$, 0, or 1 position to simulate sensor-placement changes; validation and test omit this augmentation. Pretraining validation uses mask seeds $1042+\text{batch index}$. The main model includes the auxiliary terms in \cref{app:aux}. Matching these terms and exposure budgets across J and J+R must be established from the respective configurations rather than inferred from the main-model equation.

\section{Lightweight Readout Protocol}
\label{app:readout}
These fixed-feature analyses are distinct from GPM tracking.

\paragraph{Features and splits.}
Encoders extract features in evaluation/inference mode. LaBraM uses the final three temporal tokens per channel from L12, and Re-JEPA uses the final six from Fusion; both have feature width 200. \emph{Full} denotes all downstream memory tokens, not every encoder token. \emph{Mean} is the per-frame memory-token mean, and \emph{residual} subtracts that mean from each token. Training-partition recordings are stratified by motion stage into fit/tune/held-out sets at approximately 75\%/10\%/15\%, with no recording shared across sets. The audited split contains 25,296/3,349/5,057 windows. All held-out users and user--session combinations occur in fit. Windows, expanded frame samples, and users are distinct counting units.

\paragraph{Heads and budgets.}
The multiframe MLP linearly projects flattened features per frame, concatenates the history, and applies two GELU hidden layers of width 256. Projection width is adjusted to approximately match readout parameter counts across pretrained models. The lightweight Transformer projects tokens to width 128, applies two four-head self-attention blocks with FFN width 256, and uses an MLP with two 256-dimensional hidden layers. For dynamic tasks, $q_t$ is concatenated after feature readout, not maintained as a GPM-like recurrent state. Histories include $K=1$ and $K=4$ and are reported separately.

Audited MLP runs use ten epochs, batch 256, and five seeds; Transformer runs use six epochs, batch 128, and three seeds. Both use AdamW with learning rate $3\times10^{-4}$, weight decay $10^{-4}$, and gradient clipping at 1. Results use the fixed final epoch rather than tune-selected checkpoints. The two heads do not have equal optimization-step budgets or necessarily equal capacity.

\paragraph{Targets and statistics.}
Action labels are recording-level stage labels, trained with cross-entropy and evaluated by accuracy and balanced accuracy. Pose-increment vectors and their Euclidean magnitudes are trained with MSE and evaluated with MAE in restored original units. Direction prediction uses normalized increments and cosine loss and reports directional cosine. Dynamic tasks receive the current true pose $q_t$, rather than predicting unconditionally from EMG. Target scaling uses fit data only. Relative improvement over a constant baseline is
\begin{equation}
100\left(1-\frac{\mathrm{MAE}_{\rm probe}}{\mathrm{MAE}_{\rm constant}}\right),
\end{equation}
where the constant is the fit-label mean. This baseline does not read $q_t$ and therefore does not isolate EMG's incremental contribution beyond the pose condition. Direction instability near zero increments and the absence of a $q_t$-only baseline limit interpretation.

Metrics first aggregate held-out frame samples and then readout seeds. They are neither official user-balanced tracking scores nor user-level confidence intervals. The cached data and results have been explored repeatedly; further confirmatory claims require an independent holdout.

\section{Reconstruction Constraints and Representation Collapse}
\label{app:collapse}
This appendix analyzes how reconstruction after prediction constrains completely collapsed representations and reports training without explicit variance and covariance penalties. The analysis uses the actual amplitude Smooth L1 and circular phase objectives. Complete collapse means that a representation no longer varies with sample content conditional on position and mask information. It differs from partial dimensional degeneration, redundant layers, or loss of local detail.

\subsection{Setting and Conditional Collapse}
Let $V$ denote the EMG content visible to the online branch. Let $U$ collect position, mask, valid-position, and other non-signal conditions available to the decoder. The recovery target $Y=(a,u,w)$ is determined by the original data: $a$ contains log amplitudes, $u$ contains unit-circle phases, and $w$ contains amplitude-derived phase weights. Unlike EMA representation targets, $Y$ does not change with encoder, predictor, or target-network parameters.

Write the complete predicted grid after fusion and positional restoration as
\begin{equation}
Z=H_\theta(V,U),\qquad
\widehat Y=(\widehat a,\widehat u)=D_\psi(Z,U).
\label{eq:d1}
\end{equation}
$H_\theta$ comprises the online encoder, encoder fusion, predictor, and reconstruction fusion. $D_\psi$ comprises the amplitude and phase decoders. Parameters are fixed and stochastic dropout is disabled for this analysis. The target encoder supplies only stop-gradient matching targets, not clean tokens to $D_\psi$. The decoder has no other access to $V$ or masked raw values. Sample-dependent signal information must therefore enter reconstruction through $Z$.

Define per-example reconstruction loss by
\begin{equation}
\ell_{\rm rec}(\widehat Y,Y;U)=\lambda_A\ell_A(\widehat a,a;U)
+\lambda_\phi\ell_\phi(\widehat u,u,w;U),
\qquad\lambda_A>0,\quad\lambda_\phi\geq0.
\label{eq:d2}
\end{equation}
Both terms use the target selection and within-example reductions from the method. If $\mathcal J(U)$ indexes valid target-position/frequency pairs and $n(U)=|\mathcal J(U)|>0$, then
\begin{equation}
\ell_A(\widehat a,a;U)=\frac{1}{n(U)}\sum_{j\in\mathcal J(U)}\rho(\widehat a_j-a_j),
\qquad
\rho(r)=\begin{cases}\tfrac12r^2,&|r|\leq1,\\|r|-\tfrac12,&|r|>1.\end{cases}
\label{eq:d3}
\end{equation}
All maps are assumed measurable and all loss expectations finite. Risks are compared under the same input and mask distribution. A finite first moment of log amplitudes suffices for finite Smooth L1 risk of constant amplitude predictions.

\paragraph{Definition D.1 (conditional complete collapse).}
The reconstruction-input representation is conditionally completely collapsed if a measurable function $h$ that does not read sample content satisfies
\begin{equation}
Z=h(U)\quad\text{almost surely}.
\label{eq:d4}
\end{equation}
Different channels, times, or masks may have different fixed representations. This avoids mistaking positional variation for sample information. If all online context features passed to the predictor depend only on $U$, the downstream $Z$ also satisfies \cref{eq:d4} when there is no signal bypass.

\subsection{A Reconstruction-Risk Floor for Collapsed Representations}
Define the optimal risks using non-signal conditions alone:
\begin{equation}
\mathcal R_{\rm rec}^0=\inf_g\E[\ell_{\rm rec}(g(U),Y;U)],
\qquad
\mathcal R_A^0=\inf_{g_A}\E[\ell_A(g_A(U),a;U)].
\label{eq:d5}
\end{equation}
The infima range over measurable predictions depending only on $U$, with phase outputs in the normalized range admitted by the actual loss. The optimal predictor need not be realizable by the finite decoder, and the infimum need not be attained.

\paragraph{Proposition D.1 (risk floor under complete collapse).}
Under the no-bypass structure in \cref{eq:d1}, if $Z$ satisfies Definition D.1, then every decoder $D_\psi$ satisfies
\begin{equation}
\mathcal R_{\rm rec}(\theta,\psi)
\equiv\E[\ell_{\rm rec}(D_\psi(Z,U),Y;U)]\geq\mathcal R_{\rm rec}^0.
\label{eq:d6}
\end{equation}
Its amplitude branch also satisfies $\mathcal R_A(\theta,\psi)\geq\mathcal R_A^0$.
\begin{proof}
Conditional collapse gives $D_\psi(Z,U)=D_\psi(h(U),U)$ almost surely. Set $g_\psi(U)=D_\psi(h(U),U)$. This function belongs to the prediction family in \cref{eq:d5}, so its expected loss cannot be below that family's infimum. The same argument restricted to amplitudes proves the second inequality.
\end{proof}

\paragraph{Corollary D.1 (a sufficient condition against complete collapse).}
If $\mathcal R_{\rm rec}(\theta,\psi)<\mathcal R_{\rm rec}^0$ or $\mathcal R_A(\theta,\psi)<\mathcal R_A^0$, then the representation is not conditionally completely collapsed.

A decoder relying only on positions and masks thus has a risk floor associated with the absence of sample content. Below that floor, recovery-relevant information must pass through the predicted representation. This does not require zero reconstruction error or complete recovery of the untransformed waveform.

\subsection{A Pairwise Lower Bound for the Actual Smooth L1 Objective}
The infimum in \cref{eq:d5} ranges over all measurable predictors. To obtain a computable target-only bound, draw conditionally independent $a,a'\sim P(a\mid U)$ under the same $U$ and define
\begin{equation}
\mathcal B_A=\E_U\E_{a,a'\mid U}\!\left[
\frac{1}{n(U)}\sum_{j\in\mathcal J(U)}
\rho\!\left(\frac{a_j-a'_j}{2}\right)\right].
\label{eq:d7}
\end{equation}
Target positions, frequency bins, and masks are held fixed while sample content varies. Since our masks are sampled independently of signal content, this quantity can be constructed by applying the same mask to two independent windows, while also matching any other conditions included in $U$.

\paragraph{Proposition D.2 (pairwise target bound).}
For the Smooth L1 amplitude objective in \cref{eq:d3},
\begin{equation}
\mathcal R_A^0\geq\mathcal B_A.
\label{eq:d8}
\end{equation}
If, on a positive-probability set of $U$, at least one valid target component has a non-degenerate conditional distribution, then $\mathcal B_A>0$.
\begin{proof}
For any scalar prediction $b$, convexity and even symmetry give
\begin{equation}
\begin{aligned}
\rho\!\left(\frac{a_j-a'_j}{2}\right)
&=\rho\!\left(\frac{(a_j-b)+(b-a'_j)}{2}\right)\\
&\leq\frac{\rho(a_j-b)+\rho(b-a'_j)}{2}
=\frac{\rho(a_j-b)+\rho(a'_j-b)}{2}.
\end{aligned}
\label{eq:d9}
\end{equation}
Set $b=g_A(U)_j$. Conditional on $U$, the two terms on the right have equal expectations. Averaging valid components and then $U$ yields
\[
\mathcal B_A\leq\E[\ell_A(g_A(U),a;U)].
\]
Taking the infimum over $g_A$ proves \cref{eq:d8}. If a valid component has a non-degenerate conditional distribution, independent conditional samples differ with positive probability. Since $\rho(r)>0$ exactly when $r\ne0$, the corresponding expectation is strictly positive.
\end{proof}

Combining Propositions D.1--D.2 yields a sufficient condition that requires no fitted content-free decoder:
\begin{equation}
\boxed{\mathcal R_A(\theta,\psi)<\mathcal B_A
\quad\Longrightarrow\quad Z\text{ is not conditionally completely collapsed}.}
\label{eq:d10}
\end{equation}
Nonnegative phase loss also implies $\mathcal R_{\rm rec}(\theta,\psi)\geq\lambda_A\mathcal B_A$ for collapsed representations. This condition is sufficient, not necessary: failure to beat the conservative bound does not establish collapse.

Estimation should use windows independent of model fitting, identical masking and target reductions on both sides, and account for dependence within recordings or users. Simultaneously valid confidence bounds with an upper bound on reconstruction risk below a lower bound on $\mathcal B_A$ can support \cref{eq:d10} at the corresponding confidence level. Comparing point estimates alone does not establish that statistical claim. A finite-capacity content-free baseline has true risk no lower than $\mathcal R_A^0$; beating it does not by itself exclude collapse, and its finite-sample estimate also carries uncertainty.

\subsection{Why Predictable Signal Variation Matters}
The relevant property is that visible content improves target prediction, not merely that targets vary. Consider an auxiliary squared-error analysis to characterize predictable information; it does not replace the actual Smooth L1 and circular losses.

Embed target components in a fixed-length vector $Y_A$ and let $W(U)$ be a diagonal matrix that selects valid targets and normalizes them by $1/n(U)$. Write $\|v\|_{W(U)}^2=v^{\mathsf T}W(U)v$ and assume the relevant second moments are finite. The optimal squared risks using only $U$ or both $(V,U)$ are
\begin{equation}
\mathcal R_U^{(2)}=\inf_g\E\|Y_A-g(U)\|_{W(U)}^2,
\qquad
\mathcal R_{V,U}^{(2)}=\inf_f\E\|Y_A-f(V,U)\|_{W(U)}^2.
\label{eq:d11}
\end{equation}
\paragraph{Proposition D.3 (optimal gain from visible content).}
For $\mu_U=\E[Y_A\mid U]$ and $\mu_{V,U}=\E[Y_A\mid V,U]$,
\begin{equation}
\mathcal R_U^{(2)}-\mathcal R_{V,U}^{(2)}
=\E\|\mu_{V,U}-\mu_U\|_{W(U)}^2.
\label{eq:d12}
\end{equation}
\begin{proof}
Conditional means minimize squared risk on each positively weighted component. Substitute
$Y_A-\mu_U=(Y_A-\mu_{V,U})+(\mu_{V,U}-\mu_U)$ and expand. Since $W(U)$ and $\mu_{V,U}-\mu_U$ are measurable with respect to $(V,U)$, while $\E[Y_A-\mu_{V,U}\mid V,U]=0$, the cross term has zero expectation. The remaining terms give \cref{eq:d12}.
\end{proof}
If visible EMG changes the conditional mean of target attributes, reading sample content strictly lowers optimal squared risk. When the network can realize such an improvement, a position-and-mask-only representation cannot achieve the same recovery quality. For unpredictable fluctuations conditionally independent of visible content, \cref{eq:d12} gives no strict gain. Reconstruction requires recovery-relevant information, not preservation of every observational noise component.

\subsection{Joint Optimization and Scope of the Guarantee}
For a fixed target-branch state and loss weights, write
\begin{equation}
\mathcal J(\theta,\psi)=\mathcal L_{\rm other}(\theta,\psi)
+\lambda_A\mathcal R_A(\theta,\psi)
+\lambda_\phi\mathcal R_\phi(\theta,\psi),
\label{eq:d13}
\end{equation}
where $\mathcal L_{\rm other}\geq0$ contains weighted representation matching and other retained nonnegative terms, $\lambda_A>0$, and $\lambda_\phi\geq0$. Both the main and no-VCReg configurations satisfy this nonnegativity condition.

\paragraph{Corollary D.2 (excluding collapsed global optima).}
If realizable parameters $(\theta_*,\psi_*)$ satisfy
\begin{equation}
\mathcal J(\theta_*,\psi_*)<\lambda_A\mathcal B_A,
\label{eq:d14}
\end{equation}
then conditionally completely collapsed parameters cannot globally minimize \cref{eq:d13}.
\begin{proof}
For any conditionally collapsed parameters, Propositions D.1--D.2 and nonnegativity give $\mathcal J\geq\lambda_A\mathcal R_A\geq\lambda_A\mathcal B_A$, strictly above the realizable value in \cref{eq:d14}.
\end{proof}
This sufficient condition concerns objective values, not the full EMA optimization dynamics. Collapsed stationary points can exist even when collapse has higher risk. For example, let $V=Y_A\in\{-1,1\}$ be equiprobable, $z=\theta V$, $\widehat Y_A=wz$, and the objective $\E\rho(w\theta V-Y_A)$. At $w=\theta=0$, both gradients vanish and risk is $1/2$; at $w=\theta=1$, risk is zero. The presence of reconstruction alone is therefore not a convergence guarantee from arbitrary initialization.

These conclusions concern sample information in the complete predicted path. They do not imply that every depth is independently useful, covariance is full rank, each token carries distinct local information, or recovered attributes are necessarily motion-related. Fusion may preserve information through only some layers, and decoders may use sample-varying window-level features for partial recovery. Layer quality, dimensionality, and token locality require separate measurements. Because targets are demeaned, windowed, and band-limited, recoverability here concerns specified signal attributes rather than lossless waveforms.

\subsection{Empirical Evidence without Explicit Variance and Covariance Regularization}
We train Re-JEPA w/o VCReg from scratch, changing $\lambda_{\rm var}$ from $10^{-3}$ to zero and $\lambda_{\rm cov}$ from $10^{-4}$ to zero. EMA targets, context consistency, amplitude and phase recovery, overlapping patches, physical-region masks, and AdamW weight decay remain. The term w/o VCReg denotes removal of these two explicit feature penalties only.

\begin{table}[H]
\centering
\caption{Pretraining objective weights. Both configurations are trained from scratch; $\alpha_s,\beta_s$ are the original ramp-up schedules.}
\label{tab:vcweights}
\begin{tabular}{@{}lcc@{}}
\toprule
Objective & Re-JEPA & Re-JEPA w/o VCReg\\
\midrule
Target latent & 1 & 1\\
Context consistency & $0.5\alpha_s$ & $0.5\alpha_s$\\
Amplitude recovery & 1 & 1\\
Phase recovery & 1 & 1\\
Variance & $10^{-3}\beta_s$ & 0\\
Covariance & $10^{-4}\beta_s$ & 0\\
\bottomrule
\end{tabular}
\end{table}

The no-VCReg variant uses the same 336-token grid and global batch 4608 and completes 400 epochs (E0--E399), totaling 99,200 optimizer steps. All 48 ranks passed the initial backward audit. Checkpoints contain zero variance/covariance coefficients. Raw penalty statistics can still be logged, but their weighted objective contributions are exactly zero. E399 initializes the downstream encoder, level norms, and fusion by strict loading; GPM is randomly initialized, without another downstream checkpoint.

As a descriptive loss-scale reference, the main Re-JEPA's final pretraining validation records $\mathcal L_{\rm pre}\approx0.720783$, $\mathcal L_{\rm var}\approx0.579275$, and $\mathcal L_{\rm cov}\approx1.382278$. After ramp-up, the weighted regularization contribution is approximately
\begin{equation}
10^{-3}(0.579275)+10^{-4}(1.382278)=0.0007175028,
\qquad r_{\rm VC}\approx\frac{0.0007175028}{0.720783}=0.0995\%.
\label{eq:d15}
\end{equation}
This is the objective-value fraction at one validation epoch, not a training-average fraction or a gradient-contribution ratio. Logged inputs have six-decimal precision. The no-VCReg fraction is exactly zero by its coefficients, independently of the original penalties' numerical size.

Downstream Fusion+GPM96 is frozen for four epochs and then fully fine-tuned. The variant completes E0--E24, with validation selecting E19 at step 10,460 and loss 0.282438.
\begin{table}[H]
\centering
\caption{Completed training and checkpoint selection. Validation loss is a downstream tracking objective and cannot be substituted into \cref{eq:d10}. The main model uses 48 GPUs at batch 4; w/o VCReg uses 24 GPUs at batch 4 with two-step accumulation. Both effective batches are 192, but hardware and optimization trajectories differ.}
\label{tab:vccomplete}
\begin{tabular}{@{}lcc@{}}
\toprule
Item & Re-JEPA & Re-JEPA w/o VCReg\\
\midrule
Pretraining completion & E399 / 99,200 steps & E399 / 99,200 steps\\
Variance/covariance weights & $10^{-3}/10^{-4}$ & $0/0$\\
Downstream interface & Fusion+GPM96 & Fusion+GPM96\\
Downstream epochs & E0--E20 & E0--E24\\
Frozen phase & E0--E3 & E0--E3\\
Best validation epoch & E15 & E19\\
Best validation loss & 0.280896 & 0.282438\\
Official test splits & Complete; \cref{tab:main-tracking} & Complete; \cref{tab:main-tracking}\\
\bottomrule
\end{tabular}
\end{table}

The no-VCReg test covers 45,321 windows. User-balanced angular MAEs are 6.9415$^\circ$, 9.0176$^\circ$, and 9.4653$^\circ$; landmark errors are 8.7091, 11.6106, and 12.5619 mm. User-level standard deviations are in \cref{tab:main-tracking}. These results demonstrate successful pretraining and downstream optimization without the two explicit penalties while retaining other mechanisms. Since fine-tuning updates representations, these downstream scores do not directly verify \cref{eq:d10} for pretrained features or isolate reconstruction from EMA and context consistency.

\subsection{Linking the Training Evidence to the Theoretical Condition}
The theoretical condition must be tested on E399 before downstream updates. Required quantities are $\mathcal R_A$, $\mathcal B_A$, and $\mathcal B_A-\mathcal R_A$ on the same independent evaluation distribution. These measurements have not yet been filled in; completed training and downstream test results do not replace them.

Direct representation diagnostics must distinguish variation across samples at a fixed position from variation due to position encoding. For $Z_i$ at the same channel and time location, one can report
\begin{equation}
\nu_i=\E_U[\operatorname{tr}\operatorname{Cov}(Z_i\mid U)]
\label{eq:d16}
\end{equation}
and the effective rank of that covariance. A standard deviation computed by mixing positions cannot exclude position-conditioned collapse, so mixed-token statistics are not used here as rigorous non-collapse evidence.

To analyze optimization contributions, weighted gradients $g_k=\nabla_\theta(\lambda_k\mathcal L_k)$ should be measured on the same online parameters and batch, together with their norms and directional relationships. Nonzero module gradients in the total backward pass establish participation in optimization, not the isolated contribution of reconstruction or regularization.

Propositions D.1--D.2 establish a recovery-risk lower bound for complete collapse. The no-VCReg experiment establishes trainability without explicit feature penalties. Direct verification of the risk condition and finer layer-wise or token-wise degeneration remain separate measurements.

\section{Extended Tracking Results and Source Protocols}
\label{app:extended}
This appendix retains complete tracking metrics at their recorded precision. Historical experiment records named Time-JEPA refer to the Re-JEPA main model; the naming change does not alter experiment identity. The LaBraM two-stage baseline is not a reconstruction-only ablation of Re-JEPA.

\subsection{Full Fine-tuning: Complete Metrics}
\begin{table}[H]
\centering
\caption{Complete fine-tuning metrics. LaBraM uses Stage2+GPM48 (best E18); Re-JEPA uses E399+GPM96 after four frozen epochs (best E15). Both use a provided initial pose and equal user weighting.}
\label{tab:full}
\small
\setlength{\tabcolsep}{4pt}
\begin{tabular}{@{}llrrrr@{}}
\toprule
Split & Model & Test loss $\downarrow$ & MAE ($^\circ$)$\downarrow$ & Landmark (mm)$\downarrow$ & Fingertip (mm)$\downarrow$ \\
\midrule
Cross-user & LaBraM & 0.277298 & 7.2675 & 9.3255 & 15.0456 \\
Cross-user & \textbf{Re-JEPA} & \textbf{0.260240} & \textbf{6.9093} & \textbf{8.6859} & \textbf{13.9651} \\
Cross-stage & LaBraM & 0.392081 & 10.0112 & 13.1358 & 21.7353 \\
Cross-stage & \textbf{Re-JEPA} & \textbf{0.348496} & \textbf{9.0424} & \textbf{11.6370} & \textbf{19.0677} \\
Cross-user+stage & LaBraM & 0.398250 & 10.0045 & 13.5142 & 22.3639 \\
Cross-user+stage & \textbf{Re-JEPA} & \textbf{0.368244} & \textbf{9.3577} & \textbf{12.4678} & \textbf{20.4921} \\
\bottomrule
\end{tabular}
\end{table}

\begin{table}[H]
\centering
\caption{Relative error reductions of Re-JEPA against LaBraM. Percentages describe mean differences, not statistical significance.}
\label{tab:fullgain}
\small
\setlength{\tabcolsep}{4pt}
\begin{tabular}{@{}lrrr@{}}
\toprule
Split & Angular MAE reduction & Landmark reduction & Fingertip reduction \\
\midrule
Cross-user & 4.9\% & 6.9\% & 7.2\% \\
Cross-stage & 9.7\% & 11.4\% & 12.3\% \\
Cross-user+stage & 6.5\% & 7.7\% & 8.4\% \\
\bottomrule
\end{tabular}
\end{table}

\begin{table}[H]
\centering
\caption{Comparison with published vemg2pose tracking means \citep{emg2pose}. Published means have one-decimal precision, so reductions are approximate.}
\label{tab:publishedgain}
\small
\setlength{\tabcolsep}{4pt}
\begin{tabular}{@{}lrrrrrr@{}}
\toprule
& \multicolumn{3}{c}{Angular MAE ($^\circ$)} & \multicolumn{3}{c}{Landmark error (mm)} \\
\cmidrule(lr){2-4}\cmidrule(l){5-7}
Split & vemg2pose & Re-JEPA & Reduction & vemg2pose & Re-JEPA & Reduction \\
\midrule
Cross-user & 7.7 & 6.9093 & 10.3\% & 10.3 & 8.6859 & 15.7\% \\
Cross-stage & 11.2 & 9.0424 & 19.3\% & 15.2 & 11.6370 & 23.4\% \\
Cross-user+stage & 11.0 & 9.3577 & 14.9\% & 15.4 & 12.4678 & 19.0\% \\
\bottomrule
\end{tabular}
\end{table}

All in-study results in these tables use a provided initial pose and equal user weighting. Relative gains are computed as $(\text{reference}-\text{Re-JEPA})/\text{reference}$; they are descriptive differences, not significance tests.
\FloatBarrier

\subsection{Frozen-Encoder and Layer-wise Results}
\label{app:frozen}
\begin{table}[H]
\centering
\caption{Complete frozen-encoder and layer-wise tracking results. LaBraM uses GPM48 and Re-JEPA uses GPM96. All use global batch 192, at most 20 epochs, and validation-based selection. Bold identifies the best listed frozen configuration per split.}
\label{tab:frozenfull}
\small
\setlength{\tabcolsep}{4pt}
\begin{tabular}{@{}llrrrr@{}}
\toprule
Split & Frozen representation/readout & Test loss $\downarrow$ & MAE ($^\circ$)$\downarrow$ & Landmark (mm)$\downarrow$ & Fingertip (mm)$\downarrow$ \\
\midrule
Cross-user & LaBraM L12 + GPM48 & 0.310041 & 7.8670 & 10.5992 & 17.2736 \\
Cross-user & Re-JEPA Fusion + GPM96 & 0.275201 & 7.2271 & 9.2323 & 14.9065 \\
Cross-user & \textbf{Re-JEPA L9 + GPM96} & \textbf{0.269064} & \textbf{7.1103} & \textbf{8.9948} & \textbf{14.4967} \\
Cross-user & Re-JEPA L12 + GPM96 & 0.271337 & 7.1436 & 9.1028 & 14.6658 \\
Cross-stage & LaBraM L12 + GPM48 & 0.479271 & 11.7695 & 16.2288 & 27.3855 \\
Cross-stage & Re-JEPA Fusion + GPM96 & 0.386090 & 9.8775 & 12.9227 & 21.3695 \\
Cross-stage & \textbf{Re-JEPA L9 + GPM96} & \textbf{0.370260} & \textbf{9.5357} & \textbf{12.3735} & \textbf{20.3830} \\
Cross-stage & Re-JEPA L12 + GPM96 & 0.378122 & 9.7041 & 12.6497 & 20.8753 \\
Cross-user+stage & LaBraM L12 + GPM48 & 0.472704 & 11.4415 & 16.2357 & 27.3013 \\
Cross-user+stage & Re-JEPA Fusion + GPM96 & 0.400242 & 10.0707 & 13.5671 & 22.4474 \\
Cross-user+stage & \textbf{Re-JEPA L9 + GPM96} & \textbf{0.388724} & \textbf{9.8420} & \textbf{13.1595} & \textbf{21.6948} \\
Cross-user+stage & Re-JEPA L12 + GPM96 & 0.390939 & 9.8745 & 13.2289 & 21.8597 \\
\bottomrule
\end{tabular}
\end{table}

L9 outperforms L12, and L12 outperforms Fusion, on all reported frozen metrics across the three splits. Relative to Fusion, L9 reduces angular MAE by approximately 1.6\%, 3.5\%, and 2.3\% and landmark error by 2.6\%, 4.3\%, and 3.0\%. As a descriptive cross-configuration reference, frozen L9 angular errors are also 2.2\%, 4.7\%, and 1.6\% below fine-tuned LaBraM+GPM. This does not replace a matched frozen-versus-fine-tuned comparison. The predefined Fusion interface remains the main configuration.
\FloatBarrier

\subsection{J and J+R: Complete Fine-tuning Metrics}
\begin{table}[H]
\centering
\caption{Complete objective-comparison metrics after fine-tuning. Both configurations use Fusion+GPM96 and the same tracking aggregation. Radian and degree values retain their separately recorded precision.}
\label{tab:objectivesfull}
\small
\setlength{\tabcolsep}{4pt}
\begin{tabular}{@{}llrrrrr@{}}
\toprule
Split & Objective & Test loss $\downarrow$ & MAE (rad)$\downarrow$ & MAE ($^\circ$)$\downarrow$ & Landmark (mm)$\downarrow$ & Fingertip (mm)$\downarrow$ \\
\midrule
Cross-user & J (latent-only) & 0.275972 & 0.126343 & 7.2389 & 9.2693 & 14.9629 \\
Cross-user & J+R (Re-JEPA) & 0.260240 & 0.120589 & 6.9093 & 8.6859 & 13.9651 \\
Cross-stage & J (latent-only) & 0.385553 & 0.172292 & 9.8716 & 12.8789 & 21.3261 \\
Cross-stage & J+R (Re-JEPA) & 0.348496 & 0.157819 & 9.0424 & 11.6370 & 19.0677 \\
Cross-user+stage & J (latent-only) & 0.399442 & 0.175565 & 10.0591 & 13.5214 & 22.3877 \\
Cross-user+stage & J+R (Re-JEPA) & 0.368244 & 0.163323 & 9.3577 & 12.4678 & 20.4921 \\
\bottomrule
\end{tabular}
\end{table}

J+R reduces angular, landmark, and fingertip errors relative to J by 4.6\%/8.4\%/7.0\%, 6.3\%/9.6\%/7.8\%, and 6.7\%/10.6\%/8.5\%, respectively, in cross-user/cross-stage/joint order. J uses downstream E18 (step 9937); J+R uses E15 (step 8368). Standard deviations in \cref{tab:objectives} are sample SDs of their respective per-user metrics. User identities, user counts, and per-user window counts match across configurations. A complete audit of pretraining exposure, auxiliary terms, and training budgets remains pending. No missing frozen-objective results are inferred from the fine-tuning results.
\FloatBarrier

\subsection{Complete Decoder Comparison}
\label{app:decoder}
\begin{table}[H]
\centering
\caption{Complete LSTM/GPM fine-tuning results. Bold marks the lower mean within each pretrained model. All checkpoints are validation-selected; user-level standard deviations appear in Table~\ref{tab:main-tracking}.}
\label{tab:decoderfull}
\small
\setlength{\tabcolsep}{4pt}
\begin{tabular}{@{}llrrrr@{}}
\toprule
Split & Fine-tuned configuration & Test loss $\downarrow$ & MAE ($^\circ$)$\downarrow$ & Landmark (mm)$\downarrow$ & Fingertip (mm)$\downarrow$ \\
\midrule
Cross-user & LaBraM + LSTM & 0.288776 & 7.4837 & 9.7878 & 15.8161 \\
Cross-user & \textbf{LaBraM + GPM} & \textbf{0.277298} & \textbf{7.2675} & \textbf{9.3255} & \textbf{15.0456} \\
Cross-stage & LaBraM + LSTM & 0.413124 & 10.4366 & 13.8967 & 23.0970 \\
Cross-stage & \textbf{LaBraM + GPM} & \textbf{0.392081} & \textbf{10.0112} & \textbf{13.1358} & \textbf{21.7353} \\
Cross-user+stage & LaBraM + LSTM & 0.422016 & 10.4388 & 14.4242 & 23.9825 \\
Cross-user+stage & \textbf{LaBraM + GPM} & \textbf{0.398250} & \textbf{10.0045} & \textbf{13.5142} & \textbf{22.3639} \\
Cross-user & Re-JEPA + LSTM & 0.278445 & 7.3030 & 9.3652 & 15.0983 \\
Cross-user & \textbf{Re-JEPA + GPM} & \textbf{0.260240} & \textbf{6.9093} & \textbf{8.6859} & \textbf{13.9651} \\
Cross-stage & Re-JEPA + LSTM & 0.372446 & 9.5999 & 12.4227 & 20.4895 \\
Cross-stage & \textbf{Re-JEPA + GPM} & \textbf{0.348496} & \textbf{9.0424} & \textbf{11.6370} & \textbf{19.0677} \\
Cross-user+stage & Re-JEPA + LSTM & 0.394848 & 9.9501 & 13.3871 & 22.1185 \\
Cross-user+stage & \textbf{Re-JEPA + GPM} & \textbf{0.368244} & \textbf{9.3577} & \textbf{12.4678} & \textbf{20.4921} \\
\bottomrule
\end{tabular}
\end{table}

For LaBraM, GPM reduces angular, landmark, and fingertip errors relative to LSTM by 2.9\%/4.1\%/4.2\%, 4.7\%/5.5\%/6.3\%, and 4.9\%/5.9\%/6.7\%. For complete Re-JEPA, the corresponding reductions are 5.4\%/5.8\%/6.0\%, 7.3\%/6.3\%/6.9\%, and 7.5\%/6.9\%/7.4\%. Percentages use unrounded user-balanced means.

\paragraph{Re-JEPA readout and training configurations.}
Both heads initialize from the same complete overlapping-patch E399 model, strictly loading the online encoder, L3/L6/L9/L12 norms, and hierarchical fusion. LSTM selects the final six tokens per channel and mean-pools the $16\times6$ tokens to a 200-dimensional feature. Concatenating the previous 20-dimensional pose produces the input to a conventional two-layer LSTM of hidden width 512 and 3,614,740 parameters. Its hidden state recurs within the rollout; it predicts 20-dimensional increments scaled by 0.01 before accumulation. GPM retains the same 96 tokens and uses the state-conditioned readout in \cref{sec:gpm}.

Both heads are randomly initialized, with global batch 192, up to 50 epochs, and validation-based early stopping. LSTM fine-tunes from E0 on 24 PRO6000 GPUs, batch 4 with two-step accumulation. GPM uses four frozen epochs before fine-tuning on 48 A800 GPUs, batch 4. This is therefore not a one-factor ablation of a gate or state component.

\paragraph{Checkpoint selection and test completeness.}
Re-JEPA+LSTM selects E18 (step 9937), validation loss 0.2997173667; Re-JEPA+GPM selects E15. LaBraM+LSTM selects E23 (step 12552), and LaBraM+GPM selects E18. The Re-JEPA evaluations match user identities, 20/158/20 users, 14,337/27,422/3,562 windows, and per-user window counts. Both use FP32 evaluation, 10,000-sample evaluation segments, 1790-sample context, a provided initial pose, and identical filtering of frames with failed inverse kinematics. All GPU evaluation shards completed. Historical LaBraM training specifications require further consolidation; the comparisons do not isolate the individual contributions of gates, persistent states, and attention.
\FloatBarrier

\subsection{Published Comparisons and Aggregation}
\begin{table}[H]
\centering
\caption{Published tracking means and user-level standard deviations. AE is angular MAE; LD is landmark distance. NHN values use its user aggregation, not session aggregation.}
\label{tab:publishedfull}
\small
\setlength{\tabcolsep}{4pt}
\begin{tabular}{@{}lrrrrrr@{}}
\toprule
& \multicolumn{2}{c}{Cross-user} & \multicolumn{2}{c}{Cross-stage} & \multicolumn{2}{c}{Cross-user+stage} \\
\cmidrule(lr){2-3}\cmidrule(lr){4-5}\cmidrule(l){6-7}
Source & AE ($^\circ$) & LD (mm) & AE ($^\circ$) & LD (mm) & AE ($^\circ$) & LD (mm) \\
\midrule
NHN Table 1(b) & 7.30 $\pm$ 0.88 & 9.55 $\pm$ 1.41 & 10.12 $\pm$ 1.31 & 13.59 $\pm$ 1.69 & 10.16 $\pm$ 0.99 & 14.03 $\pm$ 1.33 \\
EMG-GPT Table 4 & 8.89 $\pm$ 0.93 & 11.59 $\pm$ 1.24 & 12.42 $\pm$ 1.43 & 16.88 $\pm$ 1.80 & 12.18 $\pm$ 1.25 & 16.93 $\pm$ 1.36 \\
\bottomrule
\end{tabular}
\end{table}

NHN values come from its Table 1(b) tracking results, using the official splits, 5 s windows, and a true initial pose. Its statistics first average five seeds for each user, then compute cross-user means and SDs. NHN is an externally reported model trained with supervision from scratch, not a controlled pretraining-objective variant \citep{nhn}.

EMG-GPT uses its own scoring target indices and temporal alignment. Its Table 4 evaluates released vemg2pose checkpoints on the EMG-GPT scoring grid and explicitly does not reproduce the original benchmark protocol. We retain these tracking results as reference-only rows in \cref{tab:main-tracking}, marked $\dagger$, and do not compute matched-protocol gains against them. We consistently use its higher-precision Appendix E Table 4, not selectively choose between that table and the coarser Table 3 \citep{emggpt}.
\FloatBarrier

\subsection{Additional Training Diagnostics}
\tbd{Consolidation of archived training curves, train--validation gaps, per-user diagnostics, and repeated-seed results. Intermittently sampled training-loss means must remain distinguished from complete-epoch averages or fixed-checkpoint evaluation losses. Missing measurements are not inferred from aggregate scores.}

\section{Extended Representation Analyses}
\label{app:analysis}
\subsection{Readout Results}
The upper panel of \cref{fig:analysis} uses full downstream memory with $K=1/4$ for action classification and 40 ms increment-vector, magnitude, and direction readouts. Vector and magnitude gains are MAE improvements over the fit-label-mean constant predictor defined in \cref{app:readout}. This baseline does not read the true $q_t$, so gains do not isolate EMG's contribution beyond pose conditioning. Complete mean and seed-SD values for the 48 model/head/history/target combinations are listed below.
\begin{table}[H]
\centering
\caption{All readout results in Figure~\ref{fig:analysis}, as mean $\pm$ readout-seed SD. Action is accuracy (\%); vector/magnitude gains are MAE improvement over the fit-mean constant baseline (\%); direction is cosine. MLP and Transformer use five and three seeds, respectively. Higher is better.}
\label{tab:readoutfull}
\small
\setlength{\tabcolsep}{5pt}
\begin{tabular}{@{}rllrrrr@{}}
\toprule
$K$ & Head & Pretraining & Action & Vector gain & Magnitude gain & Direction \\
\midrule
1 & MLP & LaBraM & $18.723\pm0.121$ & $-4.500\pm0.530$ & $6.290\pm1.387$ & $0.1800\pm0.0010$ \\
1 & MLP & Latent-only JEPA & $25.188\pm0.246$ & $-0.633\pm0.633$ & $17.308\pm2.158$ & $0.2578\pm0.0015$ \\
1 & MLP & Re-JEPA & $26.869\pm0.569$ & $0.050\pm1.005$ & $20.497\pm1.490$ & $0.2816\pm0.0013$ \\
1 & Transformer & LaBraM & $25.519\pm0.145$ & $1.251\pm0.505$ & $18.168\pm1.803$ & $0.2790\pm0.0010$ \\
1 & Transformer & Latent-only JEPA & $26.394\pm0.318$ & $2.805\pm0.873$ & $21.390\pm2.126$ & $0.3020\pm0.0021$ \\
1 & Transformer & Re-JEPA & $35.542\pm0.090$ & $8.075\pm0.733$ & $30.333\pm2.042$ & $0.3741\pm0.0005$ \\
4 & MLP & LaBraM & $21.375\pm0.388$ & $-3.609\pm0.675$ & $9.993\pm0.751$ & $0.1855\pm0.0008$ \\
4 & MLP & Latent-only JEPA & $26.161\pm0.506$ & $-0.339\pm1.050$ & $13.950\pm4.152$ & $0.2567\pm0.0021$ \\
4 & MLP & Re-JEPA & $28.755\pm0.520$ & $-0.202\pm0.830$ & $17.598\pm3.459$ & $0.2874\pm0.0018$ \\
4 & Transformer & LaBraM & $26.114\pm0.196$ & $1.430\pm1.545$ & $19.364\pm2.579$ & $0.2876\pm0.0025$ \\
4 & Transformer & Latent-only JEPA & $27.677\pm0.330$ & $2.775\pm0.324$ & $23.775\pm1.324$ & $0.3057\pm0.0021$ \\
4 & Transformer & Re-JEPA & $37.178\pm0.895$ & $8.825\pm0.232$ & $30.974\pm0.977$ & $0.3782\pm0.0032$ \\
\bottomrule
\end{tabular}
\end{table}

\tbd{Additional archived mean/residual-input and longer-history results. Their reporting will retain the true-$q_t$ condition, recording-level isolation, and differences in head capacity and budget.}

\subsection{Perturbation Protocol and Responses}
\label{app:perturbation}
The lower panel of \cref{fig:analysis} uses 120 windows from the same 15 users, with paired perturbed signals for both models. The query occupies one channel over $[400,500)$ ms. Twelve neighboring physical patches satisfy $0<\Delta c^2+\Delta t^2\leq4$ on a channel-index by 100 ms grid. Every sample within a selected patch is perturbed. Donors come from another recording of the same user, are mean/SD-matched before mixing, and the mixture is rematched to the original patch's first two moments. The 25\% and 100\% settings are mixing coefficients, not measured input RMS or coverage fractions. Channel-index distance does not represent physical electrode distance.

Both models use complete native grids (144 LaBraM tokens and 336 Re-JEPA tokens). In the entry-clamped control, embeddings at locations with unchanged raw support are restored to their clean values. Layer features use a common non-affine LayerNorm, without subtracting the window mean. Let $A_u$ be the mean squared clean-to-perturbed query difference for user $u$, aggregated over windows, repetitions, and feature dimensions, and $D_u$ the natural mean squared query difference against other recordings of the same user. The reported response is
\begin{equation}
R=100\sqrt{\frac{1}{N_u}\sum_u\frac{A_u}{D_u}}.
\end{equation}
Energy ratios are computed within users, averaged with equal user weight, and then square-rooted; this is not the mean of user-wise RMS ratios. Intervals use 5000 paired user-bootstrap replicates without multiple-comparison correction. Increasing illustrated layer sizes are schematic; colors and curves encode the measured responses.
\begin{table}[H]
\centering
\caption{Relative RMS query response (\%) under 12-neighbor donor interventions. Brackets give paired user-bootstrap 95\% confidence intervals. Strength is the mixing coefficient, not coverage.}
\label{tab:responsefull}
\begin{tabular}{@{}rlll@{}}
\toprule
Strength & Layer & LaBraM & Re-JEPA \\
\midrule
25\% & L3 & 4.864 [4.409, 5.262] & 6.938 [6.535, 7.317] \\
25\% & L6 & 16.145 [15.255, 17.131] & 33.407 [29.436, 37.073] \\
25\% & L9 & 10.867 [10.191, 11.532] & 38.955 [34.432, 43.229] \\
25\% & L12 & 18.451 [17.208, 19.636] & 43.794 [39.051, 48.304] \\
100\% & L3 & 16.719 [15.851, 17.507] & 29.660 [27.413, 31.790] \\
100\% & L6 & 47.026 [44.722, 49.370] & 75.974 [69.273, 82.313] \\
100\% & L9 & 35.673 [34.471, 36.983] & 85.014 [78.490, 91.311] \\
100\% & L12 & 54.590 [52.413, 56.624] & 95.053 [88.782, 101.333] \\
\bottomrule
\end{tabular}
\end{table}

\tbd{Consolidation of archived distance, area, dose, query-self, Gaussian, amplitude/phase, and valid Fusion conditions, including absolute response, measured input RMS, zero-perturbation controls, and negative findings.}

For overlapping inputs, native-forward and query-entry-clamped controls must be reported separately. Native forward passes include direct signal effects transmitted through overlapping patches; entry clamping targets subsequent contextual propagation. They must not be combined into one response curve.

\subsection{Additional Geometry and Discrete-Code Diagnostics}
\tbd{Archived supporting analyses of full-window token similarity, mean/residual components, effective rank, interlayer CKA, and paired responses under changed/unchanged tokenizer codes remain to be consolidated. Geometry does not replace task readability, and an unchanged discrete code does not imply an unchanged Stage2 continuous representation. Correlation, group-selection effects, and negative findings must remain visible.}

\end{document}